\pdfoutput=1
\documentclass[11pt]{article}

\usepackage[preprint]{acl}

\usepackage{times}
\usepackage{latexsym}
\usepackage{amsmath}
\usepackage[T1]{fontenc}
\usepackage[utf8]{inputenc}

\usepackage{microtype}

\usepackage{inconsolata}
\usepackage{tabularx}
\usepackage{booktabs}
\usepackage{bbm}
 \usepackage{float}
\usepackage{multirow}
\usepackage{graphicx}
\usepackage{tablefootnote}
\usepackage{amsmath}
\usepackage{amssymb}
\usepackage[ruled, vlined]{algorithm2e}
\usepackage{xcolor}
\usepackage{makecell} 

\SetCommentSty{mycommfont}
\usepackage[normalem]{ulem}
\usepackage{array}
\usepackage{arydshln}
\usepackage{algpseudocode}
\usepackage{subfigure}
\usepackage{diagbox}
\usepackage{makecell}
\usepackage[most]{tcolorbox}
\usepackage{listings}

\title{Conditional [MASK] Discrete Diffusion Language Model}
\author{Hyukhun Koh\textsuperscript{1$\ast$} \hspace{1cm} Minha Jhang\textsuperscript{1$\ast$}\hspace{1cm} Dohyung Kim\textsuperscript{2} \\ {\bf Sangmook Lee\textsuperscript{2}} \hspace{1cm} \hspace{1cm}  {\bf Kyomin Jung\textsuperscript{1,2}} \\
  $^{1}$IPAI, Seoul National University
  $^{2}$Dept. of ECE, Seoul National University\\
  \texttt{\{hyukhunkoh-ai, jminha2014, kimdohyung, helmsman\}@snu.ac.kr}\\
  }

\begin{document}
\maketitle

\begin{abstract}
% Recently, large language models (LLMs) have achieved promising results in natural language processing, however, they continue to suffer from limited diversity and controllability due to the inductive biases of Auto-Regressive architecture. While there has been growing interest in combining LLMs with diffusion models, diffusion approaches themselves struggle with degenerate solutions. One possible solution is to incorporate a Masked Language Model (MLM) into a diffusion framework. However, naive integration is not effective, as MLMs and diffusion models rely on different training objectives. Therefore, in this paper, we introduce Diffusion-EAGS, a novel approach that integrates MLMs and diffusion models based on a theoretical connection of conditional Markov random fields. Experimental results demonstrate that Diffusion-EAGS outperforms existing diffusion-based baselines in conditional generation tasks, achieving both higher text quality and diversity.
% We also show that our model is capable of adapting to bilingual and low-resource settings, which are common in real-world applications.

Although auto-regressive models excel in natural language processing, they often struggle to generate diverse text and provide limited controllability. Non-auto-regressive methods could be an alternative but often produce degenerate outputs and exhibit shortcomings in conditional generation. To address these challenges, we propose \textit{Diffusion-EAGS}, a novel framework that integrates conditional masked language models into diffusion language models through the theoretical lens of a \textit{conditional Markov Random Field}. In doing so, we propose \textit{entropy-adaptive Gibbs sampling} and \textit{entropy-based noise scheduling} to counterbalance each model’s shortcomings. Experimental results show that \textit{Diffusion-EAGS} outperforms baselines and achieves the best quality-diversity tradeoff, demonstrating its effectiveness in non-autoregressive text generation.

\end{abstract}

% \footnotetext{\textsuperscript{*}Equal Contribution.}

\section{Introduction}
\begin{figure}[h]
\centering
\includegraphics[width=0.45\textwidth]{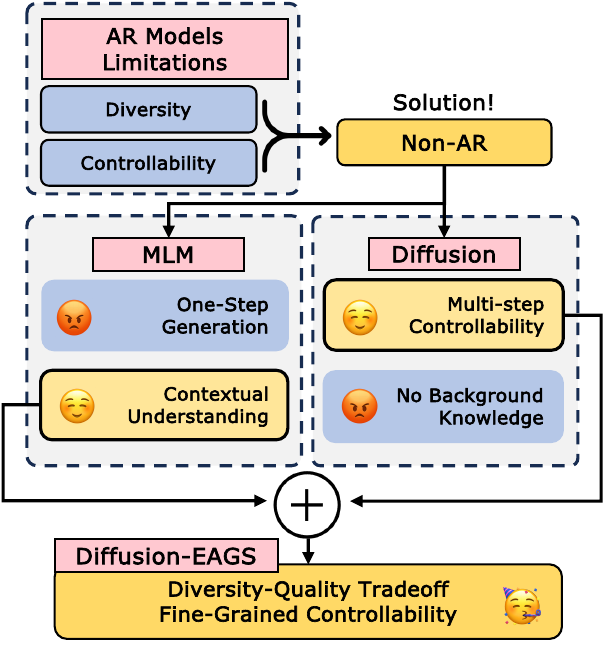}
\vspace{-0.2cm}
\caption{Overview of how our approach (Diffusion-EAGS) combines the strengths of MLM and diffusion-based models to overcome the limitations of AR models, achieving a better diversity-quality tradeoff and fine-grained controllability}
\vspace{-3mm}
\vspace{-0.3cm}
\end{figure}

Auto-Regressive Models (ARMs) have driven significant advances in NLP~\cite{achiam2023gpt,dubey2024llama,team2023gemini}, yet they still face fundamental challenges such as diversity and controllability due to the ARM's innate inductive bias. To address these challenges, a more flexible generative model is required.

Specifically, ARMs face multiple challenges: they struggle to correct mathematical reasoning errors once made~\cite{wang2025thoughtsplaceunderthinkingo1like}, and often fail to integrate external knowledge~\cite{Hudecek2023AreLA, sun-etal-2023-towards,su2024conflictbank}. These shortcomings arise from ARMs’ sequential nature, which prevents them from revising earlier steps. As a result, they cannot effectively foster diversity through temperature-based sampling alone~\cite{lee2025generatingdiversehypothesesinductive}, nor can they anticipate future requirements at earlier steps, thus undermining controllability when specific keywords must appear later~\cite{lu-etal-2022-neurologic}.

One promising alternative is non-autoregressive generation, including Conditional Masked Language Models (CMLMs)~\cite{ghazvininejad-etal-2019-mask,kasai2020non} and diffusion models. CMLMs provide strong contextual understanding but lack an effective text generation mechanism. Meanwhile, diffusion models iteratively refine text through denoising, enabling fine-grained control and increased diversity. Recent works explore direct diffusion-based generation~\cite{li2022diffusionlm,gat2024discreteflowmatching,the2024large,ye2025beyond} or hybrid approaches combining diffusion with PLMs and LLMs~\citep{lin2023textgenerationdiffusionlanguage,xiang2024diffusiondialogdiffusionmodeldiverse}. However, despite their advantages, Discrete Diffusion Language Models (DDLMs) still suffer from degeneration in conditional generation tasks~\cite{xu2025energybased}, as confirmed by our experiments.

We therefore propose \textbf{Diffusion-EAGS}, a novel approach that integrates CMLMs into DDLMs to achieve diverse, controllable, and high-quality conditional generation. However merging these methods is challenging because CMLMs generate text in one step by predicting all masked tokens, whereas diffusion models iteratively refine representations over multiple steps by introducing and removing noise. 
Our approach bridges this gap by leveraging a conditional Markov Random Field (cMRF) formulation, which enables: 
\vspace{-0.2cm}
\begin{enumerate} 
\item \textbf{Stepwise iterative generation}, overcoming the single-step limitations of CMLMs. 
\vspace{-0.2cm}
\item \textbf{Stable and diverse conditional text generation}, reducing semantic drift in DDLMs. 
\end{enumerate}
\vspace{-0.2cm}

Diffusion-EAGS achieves this through two key methodologies: 

\vspace{-0.2cm}
\begin{itemize}
\setlength{\itemsep}{0.05cm}
\item \textbf{Entropy-Adaptive Gibbs Sampling (EAGS):} A strategy that updates the most uncertain (high-entropy) tokens first at each denoising step, ensuring qualified generation. 
\item \textbf{Entropy-based Noise Scheduling (ENS):} A training approach that progressively masks tokens based on ascending order of entropy, enabling the model to learn a structured denoising process. 
\end{itemize}
\vspace{-0.2cm}

% Experimental results demonstrate that Diffusion-EAGS achieves outstanding performance compared to baselines across various conditional generation tasks. In addition, analysis involving keyword-based story generation demonstrates that our model is not only effective in generating from random mask states but also adaptable to a variety of conditioning scenarios.

% We conduct extensive experiments to validate Diffusion-EAGS on various conditional generation tasks, demonstrating significant improvements over baseline models. Moreover, keyword-based story generation experiments confirm that our model effectively generates coherent and controlled text from randomly masked sequences, making it highly adaptable to different conditioning constraints. Additionally, our approach achieves the best quality-diversity tradeoff, as evidenced by our experimental results, demonstrating that Diffusion-EAGS balances fluency and variability more effectively than existing models.

We conduct extensive experiments to validate Diffusion-EAGS on various conditional generation tasks, demonstrating significant improvements over baseline models. Our approach achieves the best quality-diversity tradeoff, demonstrating that Diffusion-EAGS balances fluency and variability more effectively than existing models. Moreover, keyword-based story generation experiments confirm that our model effectively generates coherent and controlled text from randomly masked sequences, making it highly adaptable to different conditioning constraints.

% In order to harness the strengths of diffusion, particularly its controllability and diversity, recent research has increasingly applied diffusion models to NLP, and demonstrated their promising performance~\citep{li2022diffusionlm,gong2023diffuseq,he-etal-2023-diffusionbert,yuan2023seqdiffuseq,lovelace2023latent,chen2023cheaper,he-etal-2023-diffusionbert,lou2024discrete,zhou-etal-2024-diffusion,shi2024simplified,sahoo2024simple,zheng2024masked,nie2024scalingmaskeddiffusionmodels}, and these models have been shown to be scalable~\citep{nie2024scalingmaskeddiffusionmodels}.

% Recent research has increasingly applied diffusion models to NLP, in order to leverage their strengths in controllability and diversity. These studies have demonstrated promising performance across various tasks~\citep{li2022diffusionlm, gong2023diffuseq, he-etal-2023-diffusionbert, yuan2023seqdiffuseq, lovelace2023latent, chen2023cheaper, he-etal-2023-diffusionbert, lou2024discrete, zhou-etal-2024-diffusion, shi2024simplified, sahoo2024simple, zheng2024masked,lcmteam2024largeconceptmodelslanguage,wang-etal-2024-ladic}, with diffusion-based models also proving to be highly scalable~\citep{nie2024scalingmaskeddiffusionmodels}.

\vspace{-0.1cm}
\section{Related Works}
Efforts to integrate generative flow models into sequence generation exploit the distribution shift from a source language to a target language through a series of invertible linear transformations \citep{ma2019flowseqnonautoregressiveconditionalsequence,zhang-etal-2024-languageflow}. 
However, as DDPM \citep{ho2020denoising} demonstrate the effectiveness of generating images, diffusion models have been a major topic of interest within the field of generative flow models \citep{song2021maximum, song2021scorebased}. To apply such diffusion methodologies to NLP, in order to leverage their strengths in controllability and diversity, recent studies have demonstrated promising performance across various tasks~\citep{li2022diffusionlm, gong2023diffuseq, he-etal-2023-diffusionbert, yuan2023seqdiffuseq, lovelace2023latent, chen2023cheaper, he-etal-2023-diffusionbert, lou2024discrete, zhou-etal-2024-diffusion, shi2024simplified, sahoo2024simple, zheng2024masked,the2024large,wang-etal-2024-ladic}.

% For instance, GPT o1-like models often struggle with mathematical reasoning once the first few tokens are fixed~\cite{wang2025thoughtsplaceunderthinkingo1like}, making it difficult to revise subsequent steps due to the \textit{inductive bias} inherent in autoregressive generation~\cite{wang2024chainofthoughtreasoningprompting}. In addition, in task-oriented dialogue systems, LLMs cannot properly integrate the DB search-based context, simply replicating or disregarding it~\cite{Hudecek2023AreLA,sun-etal-2023-towards}. Lastly, from a code-infilling perspective, a sufficiently large context is required to ensure optimal performance~\cite{gat2024discreteflowmatching}. 

Although Continuous Diffusion Language Models (CDLMs) such as Diffusion-LM \citep{li2022diffusionlm}, DiffuSeq-v1, v2 \citep{gong2023diffuseq,gong2023diffuseqv2}, and LD4LG~\citep{lovelace2023latent} show promising performance,  \citet{bansal2022cold} argue that such operations do not necessarily have to be governed by stochastic randomness.

 Building on this rationale, D3PM \citep{austin2023structured} propose the discrete restoration-generation approach and DiffusionBERT \cite{he2022diffusionbert} adopt pre-trained language models (PLMs) to DDLM. SEDD~\cite{lou2024discrete} propose score entropy inspired by MLM loss, and outperform existing CDLMs. Recent works by \citet{shi2024simplified} and \citet{sahoo2024simple} extend this idea and obtain better empirical results. \citet{zheng2024masked} further enhance discrete diffusion models by correcting the numerical precision error in SEDD-based models. These research make an improvement on the open ended generation task. Furthermore, \citet{venkatraman2024amortizing} use SEDD as text infilling, and \citet{nie2024scalingmaskeddiffusionmodels} demonstrate that DDLMs are scalable.

\section{MLM \& DDLM : D-cMRF}
\label{sec:theory}
Pre-trained MLMs offer rich, context-aware representations through one-pass masked prediction, whereas DDLMs iteratively refine text via stepwise denoising to enhance control and diversity. Combining these approaches can overcome MLMs’ one-pass limitations and DDLMs’ degeneration in conditional generation. However, their integration is challenging because DDLMs require iterative updates while MLMs predict all masked tokens simultaneously. To bridge this gap, we propose Diffusion-based Constrained Markov Random Fields (D-cMRF), a framework that integrates a discrete diffusion process into MLM sequence generation. By leveraging an entropy-based sampling strategy to selectively update high-uncertainty tokens at each step, D-cMRF achieves a principled reduction in sequence energy, leading to stable and coherent generation.

\vspace{-0.1cm}
\subsection{MLM as cMRF}
\label{sec:BERT_CMRF}
Inspired by the traditional approaches of \citet{wang-cho-2019-bert} and \citet{goyal2022exposingimplicitenergynetworks}, which model MLMs as Markov Random Fields (MRFs) and energy-based models (EBMs), respectively, we reinterpret MLM as a conditional MRF (cMRF) model and employ it as a denoising function at each diffusion step.

Let \( X = (x_1, x_2, \dots, x_L) \) be a sequence of discrete variables from a vocabulary \( V \), with \( Y \) representing observed conditions. The sequence probability follows an energy-based MRF formulation:

\begin{small}
\begin{equation}
    P_{\theta}(X ; Y) = \frac{\exp(-E_{\theta}(X; Y))}{Z(Y, \theta)}
\end{equation}
\end{small}
\vspace{-0.3cm}

where \( E_{\theta}(X; Y) \) is the \textbf{energy function} parameterized using MLM-based embeddings, ~$\theta$ denotes parameterization of MLM, and \( Z(Y, \theta) \) is the \textbf{partition function} for ensuring proper normalization. Then the total sequence energy is defined as:

\begin{small}
\begin{equation}
    E_{\theta}(X; Y) = -\sum_{l=1}^{L} \log \phi_l(X; Y)
\end{equation}
\end{small}
\vspace{-0.3cm}

where \textbf{log-potential function} $\log \phi_l(X; Y)$ is :

\begin{small}
\begin{equation}
    \log \phi_l(X; Y) = 1h(x_l)^T f_{\theta}(X \backslash \{x_l\}; Y)
\end{equation}
\end{small}
\vspace{-0.3cm}

where \( l \) is a token position in the sequence, \( 1h(x_l) \) is the one-hot encoding of token \( x_l \), and \( f_{\theta}(X \backslash \{x_l\}; Y) \) represents the MLM logit output conditioned on the sequence. 

% Unlike prior works \cite{wang-cho-2019-bert,goyal2022exposingimplicitenergynetworks}, 
% A detailed comparison between our work and previous studies can be found in the appendix~\ref{}.

\vspace{-0.1cm}
\subsection{DDLM with Entropy-based Denoising}
\label{sec:DDLM_egs}
Determining how to perform sampling with such a simple cMRF presents a separate challenge. In particular, one can use techniques such as Gibbs sampling as long as the energy space remains unchanged, but we cannot guarantee that this energy space is stable in general~\cite{goyal2022exposingimplicitenergynetworks}. The necessity of generating sequences in cMRF based on energy update is in Appendix~\ref{appx:need_engergy}.
Hence, a natural research question arises: 
\emph{``How should we sample and update the energy?''} 

The training process of diffusion models (both forward and backward) conceptually represents the entire distribution as a product of local conditional distributions across time steps. 
Hence, diffusion models share a probabilistic graphical structure 
with MRF, enabling MLM to be integrated within DDLM framework.

Therefore, in this subsection, we describe how to update the energy and perform sampling under the DDLM framework using \(P_{\theta}(X ; Y)\). Specifically, we integrate \(P_{\theta}(X ; Y)\) into each diffusion step as a denoising function, employing an entropy-based denoising matrix $Q$ in Section~\ref{sec:forward process}. We first define the entropy of each token:

\begin{small}
\begin{equation}
H_i(X^{(t)}) = - \sum_{x' \in V} p_{\theta}(x'_i ; X^{(t)}) 
\log p_{\theta}(x'_i ; X^{(t)})
\end{equation}
\end{small}
\vspace{-0.3cm}

where \( p_{\theta}(x'_i ; X^{(t)}) \) is the softmax probability of token $x'_i$ at position $i$ in sequence $X^{(t)}$, and $t$ denotes the diffusion timestep. 
% from the MLM, $i$ for a position of sequence, and $t$ for a timestep in the diffusion framework. We then select high-entropy positions for updating:
We then select high-entropy positions for updating:

\begin{small}
\begin{equation}
\label{eq:entropy_idxs}
    M_t = \{ i \mid H_i(X^{(t)}) \geq \tau_t \}
\end{equation}
\end{small}
\vspace{-0.3cm}

where \( \tau_t \) is a dynamically adjusted entropy threshold. This ensures that updates occur at positions where the model has the highest uncertainty. Subsequently, we sample the next-step sequence from \(P_{\theta}(X^{(t)} ; Y)\) at the suggested positions. We perform this selection process at every diffusion step, which corresponds to updating the energy, different from existing research~\cite{wang-cho-2019-bert,goyal2022exposingimplicitenergynetworks}.

\subsection{D-cMRF}
% By combining DDLM with cMRF, our approach enables a theoretically grounded generation process from the perspective of BERT, and from the standpoint of diffusion, the diffusion training process is naturally aligned with BERT's MLM objective as discussed in Section~\ref{sec:BERT_CMRF} and~\ref{sec:DDLM_egs}. Specifically, our D-cMRF guarantees monotonic energy reduction during generation, ensuring stable sequence reconstruction.
By combining DDLM with cMRF, our approach enables a theoretically grounded generation process from the perspective of MLM. Moreover, from the diffusion standpoint, the training process naturally aligns with MLM objective, as discussed in Section~\ref{sec:BERT_CMRF} and Section~\ref{sec:DDLM_egs}. Specifically, our D-cMRF guarantees energy reduction during generation, ensuring stable sequence reconstruction.
\begin{figure*}[h]
    \centering
    \includegraphics[width=1\textwidth]{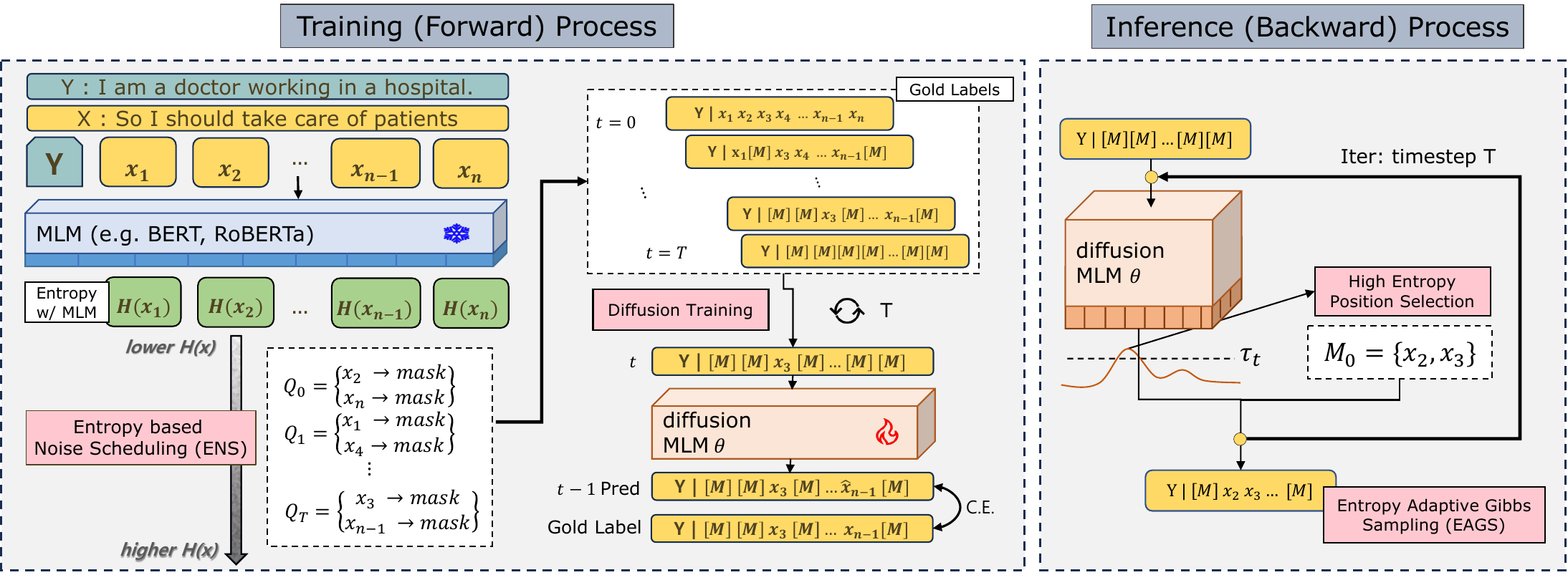}
    \vspace{-0.7cm}
    \caption{Overview of the training (forward) and inference (backward) processes in Diffusion-EAGS. \textbf{Training (left):} Entropy-based Noise Scheduling (ENS) determines which tokens in the masked sequence, denoted by $[M]$, should be denoised at each timestep based on the position entropy $H(x_i)$. These tokens are then generated using the diffusion model with parameters $\theta$, and the loss is computed using a cross-entropy (C.E.) diffusion loss. \textbf{Inference (right):} Starting from a fully masked sequence conditioned on $Y$, Entropy-Adaptive Gibbs Sampling (EAGS) iteratively refines the sequence by focusing on high-entropy tokens, denoted as $M_t$, based on a threshold $\tau_t$, yielding stable and coherent text generation.}
    \vspace{-0.5cm}
    \label{fig:overview}
\end{figure*}
\paragraph{Step 1: Expected Energy at Diffusion Step \( t \)}

At diffusion step \( t \), the expected sequence energy is defined as follows:

\begin{small}
\begin{equation}
    \mathbb{E}_{X^{(t)} \sim q} \left[ E_{\theta}(X^{(t)}; Y) \right] = \sum_{X^{(t)}} q(X^{(t)}) E_{\theta}(X^{(t)}; Y)
\end{equation}
\end{small}
where $q(\cdot)$ denotes the probability distribution from which $X^{(t)}$ is sampled. Since high-entropy tokens are selected for replacement, the total sequence energy can be decomposed as follows:

\begin{small}
\begin{equation}
\begin{aligned}
\mathbb{E} \bigl[E_{\theta}(X^{(t)}; Y)\bigr] 
&= \sum_{i \in M_t} \mathbb{E} \bigl[ E_{\theta}(x_i^{(t)}; X^{(t-1)}, Y) \bigr] \\
&\quad + \sum_{i \notin M_t} E_{\theta}\bigl(x_i^{(t-1)}; Y\bigr).
\end{aligned}
\end{equation}
\end{small}

\paragraph{Step 2: Energy Reduction via Denoising}

Since masked tokens are replaced with lower-energy candidates at each step, we expect a general trend of \textbf{energy reduction}. However, due to the stochastic nature of sampling, local fluctuations in energy may occur. Over multiple diffusion steps, the entropy-based selection mechanism ensures a net decrease in sequence energy.

\begin{small}
\begin{equation}
    \mathbb{E} \left[ E_{\theta}(x_i^{(t)}; X^{(t-1)}, Y) \right] \leq E_{\theta}(x_i^{(t)}; X^{(t)}, Y)
\end{equation}
\end{small}

Applying this property across all updated tokens \( i \in M_t \), we obtain:

\begin{small}
\begin{equation}
    E_{\theta}(X^{(t-1)}; Y) \leq E_{\theta}(X^{(t)}; Y)
\end{equation}
\end{small}

\paragraph{Step 3: Convergence to Low-Energy States}

Summing over all diffusion steps \( T \):

\begin{small}
\begin{equation}
    E_{\theta}(X^{(0)}; Y) \leq E_{\theta}(X^{(T)}; Y)
\end{equation}
\end{small}
where \( X^{(T)} \) is the fully masked sequence with maximum entropy, and \( X^{(0)} \) is the final reconstructed sequence. Since the token space is discrete and energy is derived from a sum of bounded logits, \( E_{\theta}(X; Y) \) is \textbf{lower-bounded} by a finite minimum energy state. While stochastic sampling may introduce fluctuations, the diffusion process ensures \textbf{progressive energy minimization}, leading to an approximate low-energy state.

\subsubsection{D-cMRF Guarantees}
The above proof establishes that our method satisfies the following properties:
\vspace{-0.2cm}
\begin{itemize}
    \item \textbf{Progressive Energy Reduction:} The energy function exhibits an overall decrease, leading to more stable sequence generation. This trend is supported by empirical results in Appendix~\ref{appendix:entropy_flow}.
    \vspace{-0.2cm}
    \item \textbf{Stable Convergence:} Since the energy function is lower-bounded and sequence length is finite, the generation process is expected to reach a structured, low-entropy state.
\end{itemize}
\vspace{-0.2cm}
These properties explain the improved performance of Diffusion-EAGS compared to traditional diffusion models, as shown in \S Section~\ref{sec:results}. Notably, the ablation study in Table~\ref{tab:ablation} demonstrates that removing EAGS leads to a significant drop in performance, highlighting its importance in guiding stable generation.

\section{Diffusion-EAGS}
Our approach, Diffusion-EAGS, leverages two key components—Entropy-Adaptive Gibbs Sampling (EAGS) and Entropy-based Noise Scheduling (ENS)—rooted in the theory of Section~\ref{sec:theory}. As shown in Figure~\ref{fig:overview}, during training, ENS selectively masks tokens based on their certainty, while during generation, EAGS iteratively refines a fully masked sequence by updating high-uncertainty tokens. This stepwise refinement yields balanced improvements in text quality and diversity.

\subsection{Inference Process: Entropy-Adaptive Gibbs Sampling}

% Building on the assumption that conditional generation in BERT can be modeled as a CRF, we propose a method to fully leverage the stepwise generation process inherent to diffusion models.

% In discrete diffusion models for language generation, each step involves denoising some tokens \cite{he2022diffusionbert}, which aligns with the Gibbs sampling approach in CRF. Specifically, we introduce an entropy-based adaptive Gibbs sampling method that maximizes the utility of the stepwise generation process.

% Since the diffusion model learns to how to generate the language in a sequential manner, which allows to train the basic language distribution, enabling it to emulate the potential for the sequence \( X \). 
% Additionally, as the condition \( Y \) is observed, the semantic and syntactic structures are restricted.

% Therefore, given a sequence with masked tokens \([Y, [\text{MASK}], [\text{MASK}], \ldots ]\), we model actual potential function based on the language distribution with the potential function provided by the diffusion-trained model. Using the predicted potential functions, we can perform adaptive Gibbs sampling to improve generation performance.

% \vspace{-0.4cm}

As discussed in Section~\ref{sec:DDLM_egs}, MLM can be interpreted as cMRF, which is used as \(p_\theta\) in the sampling process of DDLM with $M_t$. In particular, \(M_t\) is not only associated with energy updates but also serves as a solution to the MLM's difficulty in selecting the next tokens to denoise, as shown in Appendix ~\ref{appdx:entropy_guide}. Henceforth, we designate this strategy as \textit{Entropy-Adaptive Gibbs Sampling (EAGS)}.

In EAGS, $M_t$ is constructed  by ranking tokens in descending order of entropy, thereby prioritizing the least informative parts of the sequence. EAGS facilitates the creation of more structured sequences by leveraging the syntactic context that has already been established. The process of determining the denoising schedule is shown in Appendix~\ref{appx:entropy_selection_criteria}.

Our approach for T step-generation process can be formalized as follows:
\vspace{-0.2cm}
\begin{enumerate}
  \setlength\itemsep{0em}
    \item \textbf{Entropy Calculation}: Compute the entropy $H_i(X^{(t)})$ for each variable \( x_i \).
    \item \textbf{Variable Selection}: Obtain $M_t$ for sampling
    
    % \item \textbf{Variable Selection}: Select the variable \( x_{i^*} \) with the highest entropy for sampling
    
    % \quad \quad \quad \text{\footnotesize $l^* = \arg\max_{l} H(x_l \mid  Y, f_\theta)$} 
    \item \textbf{Sampling}: Sample \( x_{i^*} \) from its conditional distribution \( p_{\theta}(x_{i^*} \mid X^{(t)},Y) \), where \( i^* \in M_t \).

    \item \textbf{Update}: Update the conditional distributions and entropy for the affected variables.

    \item \textbf{Iteration}: Repeat Steps 1 through 4 until $t=T$, where $T$ is the total number of timestep.
\end{enumerate}
\vspace{-0.2cm}

The detailed algorithm of EAGS is in Appendix Algorithm~\ref{alg:EAGS}.

\subsection{Training Process: Entropy-based Noise Scheduling}
\label{sec:forward process}
To improve the effectiveness of EAGS during generation, we simulate a similar process during training. Therefore, we schedule the forward process of diffusion training based on the entropy $H_i(X^{(t)})$ of position $x_i$ with the input sequence \([Y|X^{(t)}]\) at sampled timestep $t$. During training, $H_i(X^{(t)})$ is calculated by pre-trained MLM. Assuming the diffusion process progresses over \( T \) steps, we mask the \( L/T \) number of positions with the lowest entropy from the set \(\{x_1, \ldots, x_L\}\) at each step \( t \), where $L$ is the sequence length. This selection process is used to determine \(\tau_t\) in Equation~\ref{eq:entropy_idxs}. The masking process at step \( t \) in position \(i\) is described by the denoising matrix \( Q_{ti} \).
\vspace{-0.3cm}
\[ \text{\footnotesize $
Q_{ti} = \begin{bmatrix}
q_{11} & 0 & \cdots & 0 & q_{1,M} \\
0 & q_{22} & \cdots & 0 & q_{2,M} \\
\vdots & \vdots & \ddots & \vdots & \vdots \\
0 & 0 & \cdots & q_{M-1,M-1} & q_{M-1,M} \\
0 & 0 & \cdots & 0 & q_{MM} \\
\end{bmatrix} $}
\]
Here, $q_{1,M}$ denotes the transition probability from the vocab index corresponding to token 1 to the [MASK] token and $q_{mn}$ is defined as:

\vspace{-0.4cm}
\fontsize{8pt}{9pt}\selectfont
\[
q_{mn} = 
\begin{cases} 
q_{mm} = 1 & \text{if } x_i \notin \text{MIN}([H_1(x_1), \cdots ,H_L(x_L)],\frac{L}{T}) \\
q_{mM} = 1 & \text{if } x_i \in \text{MIN}([H_1(x_1), \cdots ,H_L(x_L)],\frac{L}{T}) \\
0 & \text{otherwise}
\end{cases}
\]
\vspace{-0.4cm}
\normalsize

Henceforth, we designate this strategy as \textit{Entropy-based Noise Sampling (ENS)}. ENS masks lower entropy tokens first, thereby learning to progressively generate sequences. This ensures that the forward process in diffusion training closely mirrors the generation process, thereby enhancing the effectiveness of EAGS in language generation. The detailed algorithm of ENS is in Appendix Algorithm~\ref{alg:ENS}.

% \subsection{Hidden Entropy State Tracking}

% Throughout the noising and denoising phases, PLMs often face challenges in identifying which tokens should be denoised. To address this challenge, we incorporate an additional module named Absorbing State Space Tracking (ASST) module, which operates as an absorbing state pointer model. When the intermediate representation $z_{t+1}$ goes into ASST, it pinpoints the most determinant position among tokens in $z_{t}$ based on the entropy values. ASST enhances the efficiency of the generation process by imitating the process of constructing sentences.

\subsection{Diffusion Loss with Cross Entropy}

\newcommand{\algrule}[1][.2pt]{\par\vskip.5\baselineskip\hrule height #1\par\vskip.5\baselineskip}

Distinct from the prevailing methodologies in diffusion models~\cite{ho2020denoising,austin2023structured}, we do not employ the PLM parameterization \( \widetilde{p}_{\theta}(\widetilde{z}_0 | z_t, t) \), which preserves the original semantic embedding spaces during the training phase as we empirically find that such method restricts the diversity of generated responses. We follow the traditional diffusion loss~\citep{NEURIPS2020_4c5bcfec}, changing Mean Squared Error with Cross Entropy Loss.

\section{Experiments}
\vspace{-0.1cm}
% In this section, we outline the experimental configurations. Our goal is to validate the efficiency of our model in dataset-guided generation.

\subsection{Tasks \& Details}
We conduct experiments on various conditional generation datasets. Detailed explanation of the conditional generation datasets are in Appendix~\ref{appendix:datasets}. In particular, we focus on two datasets that significantly differ in their level of conditional constraints: RocStories~\citep{mostafazadeh2016corpus}, which is relatively open-ended, and Paradetox~\citep{logacheva-etal-2022-paradetox}, which imposes the strongest conditional constraints. We select the conditional dataset that GPT-2 faces in generating sentences of appropriate length under specified conditional constraints. The maximum lengths of Paradetox and RocStories is set to 64, based on data statistics, and other details are in Appendix~\ref{appendix:exp}. We test 20 conditions with 5 outputs in total 100, which is not used for training. The number of steps of our model is configured to 5 with a naive categorical sampling with a sample size of 20 and select final 5 samples based on Perplexity score. We use 1 A100 GPU with the batch size as 256.

\vspace{-0.1cm}
\subsection{Baselines}
% We employ RoBERTa-base~\citep{liu2020roberta} as MLM with learning rate 5e-4. Next, we compare Diffusion-EAGS with four categories of baselines with the model similar to the size of \textit{RoBERTa-base}: Auto-regressive Models (ARMs), Conditional Masked Language Models (CMLMs), Continuous Diffusion Language Mdoels (CDLMs), and Discrete Diffusion Language Models (DDLMs). Note that our primary goal is to investigate the architecture's inherent capabilities, any baseline approach in the direction of scalability or attempting to circumvent the architecture’s limitations goes beyond the scope of our research.
We employ RoBERTa-base~\citep{liu2020roberta} as MLM with learning rate 5e-4. Next, we compare Diffusion-EAGS with four categories of baselines of similar size to \textit{RoBERTa-base}: Auto-regressive Models (ARMs), Conditional Masked Language Models (CMLMs), Continuous Diffusion Language Models (CDLMs), and Discrete Diffusion Language Models (DDLMs). Note that our primary goal is to investigate the architecture's capabilities; any baseline approach in the direction of scalability or bypassing the architecture’s limitations goes beyond our research scope.

For \textbf{ARMs}~\citep{vaswani2023attention}, we employ GPT-2 \citep{radford2019language} and gpt-3.5-turbo\footnote{\url{https://platform.openai.com/docs/models/gpt-3-5}} with four-shot prompt. More experimental details of GPT-3.5 can be found in Appendix~\ref{appedix_gpt}. For \textbf{CMLMs}, we utilize CMLM-Mask-Predict \citep{ghazvininejad-etal-2019-mask} and DisCo-Easy-First~\citep{kasai2020non}, which are transformer-based NAR models. For \textbf{CDLMs}, our baseline includes DiffuSeq~\citep{gong2023diffuseq},  LD4LG~\citep{lovelace2023latent}, and DINOISER~\citep{ye2024dinoiser}. DiffuSeq and DINOISER is designed for sequence-to-sequence applications, and LD4LG adopt \textit{BART} as denoising init point. For \textbf{DDLMs}, we utilize DiffusionBERT~\citep{he2022diffusionbert}, applying \textit{BERT} into DDLMs, and SEDD~\citep{lou2024discrete}, a powerful open-ended generation DDLM. For SEDD, we download the pre-trained version and fine-tune it. More details are in Appendix~\ref{appdx:metrics_hyperparams}.

\vspace{-0.1cm}
\subsection{Metrics}
\paragraph{Quality metrics}
In addition to our theoretically guided methods, we evaluate performance using multiple metrics. Specifically, we use Perplexity (PPL) based-on GPT-2 Large and GPT-2 XL as an automated metric, MAUVE~\cite{pillutla2021mauve} to assess style consistency between the training data and generated outputs, SOME \citep{yoshimura-etal-2020-reference} to score the grammar, Mean Opinion Score (MOS) from human evaluations to gauge text quality, and LLM score such as LLM-c~\citep{lin-chen-2023-llm} to measure the plausibility of the narratives as a sub-metric.

\paragraph{Diversity Metrics} 
Following our quality assessment, we evaluate diversity through three different measures: an automatic frequency-based metric n-gram Vendi Score(VS n-gram)~\cite{friedman2023vendi}, a neural network–based semantic metric SimCSE Vendi Score (VS emb), and a human evaluation score MOS. The detailed descriptions of metrics are provided in Appendix~\ref{appdx:metrics_hyperparams} and~\ref{appedx:llmeval_prompt}.

\begin{table}[]
\centering
\renewcommand{\arraystretch}{1.2}
\resizebox{\linewidth}{!}{%
\begin{tabular}{ll ccc} \toprule
        \multirow{2}{*}{Model}  & \multicolumn{3}{r}{Text Quality} \\
        \cmidrule(lr){3-5}
                      & Step & PPL $\downarrow$          & MAUVE $\uparrow$        & MOS $\uparrow$      \\ \hline
\multicolumn{5}{l}{\textit{AR model}} \\
GPT-2                      & 1    & 389.1       & 0.503         & 0.83                \\ 
GPT-3.5 w/ 4-shot                      & 1    & 104.375      & 0.175         & 1                \\ \hline
\multicolumn{5}{l}{\textit{CMLMs}} \\
CMLM w/ Mask-Predict &  10& 669.9 & 0.0234 &  - \\
DisCo w/ Easy-First & 10& 716.1 & 0.0344 & - \\ \hline
\multicolumn{5}{l}{\textit{Diffusion models}} \\
DiffusionBERT              &  2000    & 775.9       & 0.737         &  0.88        \\ 
DiffuSeq                   &  2000    & $\geq1k$      & 0.683         &  -                \\
SEDD                 &  1024    & $\geq1k$       & NA        &  -        \\ 
LD4LG	                   &  2000    & 579.9       & 0.556         &   0.91        \\
DINOISER                   &  20      & 124.8       & 0.255         &  0.91       \\ \hline
\textbf{Diffusion-EAGS}      &  5      & 109.3 & 0.811         & 0.97       \\ \bottomrule
\end{tabular}}
\vspace*{-0.2cm}
\caption{Text quality of conditional generation outputs. We report Perplexity (PPL) for sentence fluency, MAUVE for condition alignment, and Mean Opinion Score (MOS) for semantic coherence. Models with PPL exceeding 600 were excluded from human evaluation.}
\vspace{-0.1cm}
\label{tab:quality}
\end{table}
 
\begin{table}[t]
\centering
\renewcommand{\arraystretch}{1.2}
\vspace*{-0.1cm}
\resizebox{\linewidth}{!}{%
\begin{tabular}{l ccc cc}
\toprule
\multirow{1}{*}{Model}  & \multicolumn{3}{c}{Text Quality} & \multicolumn{2}{c}{Diversity} \\
\cmidrule(lr){2-4} \cmidrule(lr){5-6}
                      & PPL $\downarrow$ & SOME $\uparrow$        & LLM-c $\uparrow$   &    VS(ngram) $\uparrow$      & self-bleu $\downarrow$  \\ \hline
Original Data              & 100.6             & 0.895                 & 1                  &                            &                  \\ \hline
GPT-2                      & 88.5             & \textbf{0.856}      & \textbf{0.88}         & 4.722                     & 0.124                  \\ 
DiffusionBERT              & 318.2         & 0.783                  & 0.72                  & 4.735                     & 0.088             \\
SEDD              & 273.2         & 0.827                  & 0.59
                        & \textbf{4.859}          & \textbf{0.044}      \\
\hline
\textbf{Diffusion-EAGS}         & \textbf{67.3}        & 0.844          & 0.87                   & 4.837            & 0.058        \\
\bottomrule
\end{tabular}}
\vspace*{-0.2cm}
\caption{Results on the open-ended RocStories (ROC) dataset. We report perplexity (PPL) for fluency, SOME and LLM-c for text quality, and both VS(\textit{ngram}) and self-BLEU for diversity.}
\vspace{-0.4cm}
\label{tab:roc}
\end{table}

\vspace{-0.1cm}
\section{Results}
\label{sec:results}

In Tables~\ref{tab:quality}, \ref{tab:roc}, and \ref{tab:diversity}, our model consistently demonstrates strong text quality and diversity compared to various baselines across a wide range of conditional generation tasks.

\paragraph{Text Quality.} Table~\ref{tab:quality} shows that our model achieves notable improvements in perplexity (PPL) and obtains high MAUVE and MOS scores, indicating that the generated texts are both fluent and coherent. 
Although GPT-3.5-turbo is capable of generating high-quality text, the MAUVE metric indicates that few-shot prompts alone are insufficient for accurately replicating the dataset’s inherent characteristics. On the other hand, CMLMs, DiffuSeq, and DINOISER can handle conditional constraints but sometimes struggle with semantic drift or high PPL. In contrast, Diffusion-EAGS achieves both lower PPL and strong human evaluation scores (MOS), suggesting that it effectively balances condition satisfaction with text quality. Table~\ref{tab:roc} further demonstrates our model’s capability on the open-ended RocStories dataset. Even with minimal constraints, Diffusion-EAGS maintains competitive scores compared to GPT-2, demonstrating its robustness in narrative generation.
\begin{table}[h]

\centering
\renewcommand{\arraystretch}{1.2}
\resizebox{\linewidth}{!}{%
\begin{tabular}{ll ccc} \toprule
        \multirow{2}{*}{Model}  & \multicolumn{3}{r}{Diversity} \\
        \cmidrule(lr){3-5}
                      & Step & VS(ngram) $\uparrow$    & VS(emb) $\uparrow$      & MOS $\uparrow$      \\ \hline
\multicolumn{5}{l}{\textit{AR model}} \\
GPT-2                      & 1    & 3.925         & 2.640         &  2.65        \\ 
GPT-3.5 w/ 4-shot          & 1    & 3.098         & 1.915         &  2.2        \\ \hline

\multicolumn{5}{l}{\textit{CMLMs}} \\
CMLM w/ Mask-Predict &  10   & \textcolor{gray}{1.000} & \textcolor{gray}{1.000} & -  \\
DisCo w/ Easy-First & 10   & \textcolor{gray}{1.000} & \textcolor{gray}{1.000} & -  \\
 \hline
\multicolumn{5}{l}{\textit{Diffusion models}} \\
DiffusionBERT              &  2000  & 3.101 & 2.058 & 2  \\
DiffuSeq                   &  2000  & \textcolor{gray}{2.059} & \textcolor{gray}{1.465} & -  \\
SEDD                 &  1024  & \textcolor{gray}{4.746} & \textcolor{gray}{4.063} & -  \\ 
LD4LG	                   &  2000  & 1.914         & 1.425         & 1         \\
DINOISER                   &  20    & 2.287         & 2.174         &  1        \\ \hline
\textbf{Diffusion-EAGS}      &  5     & \textbf{4.417} & \textbf{3.311} & \textbf{4.6}  \\ \bottomrule
\end{tabular}}
\vspace*{-0.3cm}
\caption{Diversity evaluation for generated outputs. We report the n-gram-based Vendi Score (VS(\textit{ngram})), the embedding-based Vendi Score (VS(\textit{emb})), and a Mean Opinion Score (MOS) for diversity. Higher values indicate greater diversity.}
\vspace*{-0.6cm}
\label{tab:diversity}
\end{table}

\paragraph{Diversity.} Diffusion-EAGS excels at generating diverse outputs. As illustrated in Table~\ref{tab:diversity}, our model consistently excels in both n-gram and embedding-based diversity metrics (VS(ngram) and VS(emb)), surpassing other baselines and even larger LLMs. The model’s higher MOS for diversity further indicates that humans also perceive its outputs to be more varied and engaging. In line with these observations, we conduct additional analyses (Appendix~\ref{appdx:diversity_saturation}) including the comparison ours with large LLMs, where our approach produces a wider range of coherent yet distinct responses. These findings underscore the effectiveness of our entropy-adaptive sampling strategy in avoiding repetitive outputs and semantic collapse, thereby delivering a superior quality-diversity trade-off.

Overall, \textbf{Diffusion-EAGS} consistently demonstrates \textit{strong performance across diverse conditional generation tasks}, combining low perplexity and high human evaluation scores with the ability to generate richly varied text. Detailed results are in Appendix~\ref{appdx:results} and examples are in Appendix~\ref{appdx:generated_examples}.

% \section{Example: Image Generation}
% \input{06_image_generation}
\section{Analysis}

\subsection{Quality-Diversity Tradeoff}
Balancing \textit{quality} and \textit{diversity} is a fundamental challenge in text generation. AR models typically achieve high fluency but suffer from low diversity, while non-autoregressive models, such as CMLMs and diffusion models, often struggle to generate coherent outputs. Our proposed \textbf{Diffusion-EAGS} effectively balances these factors by leveraging a structured diffusion process.

\begin{figure}[]
\centering
\vspace*{-0.4cm}
\includegraphics[width=0.5\textwidth]{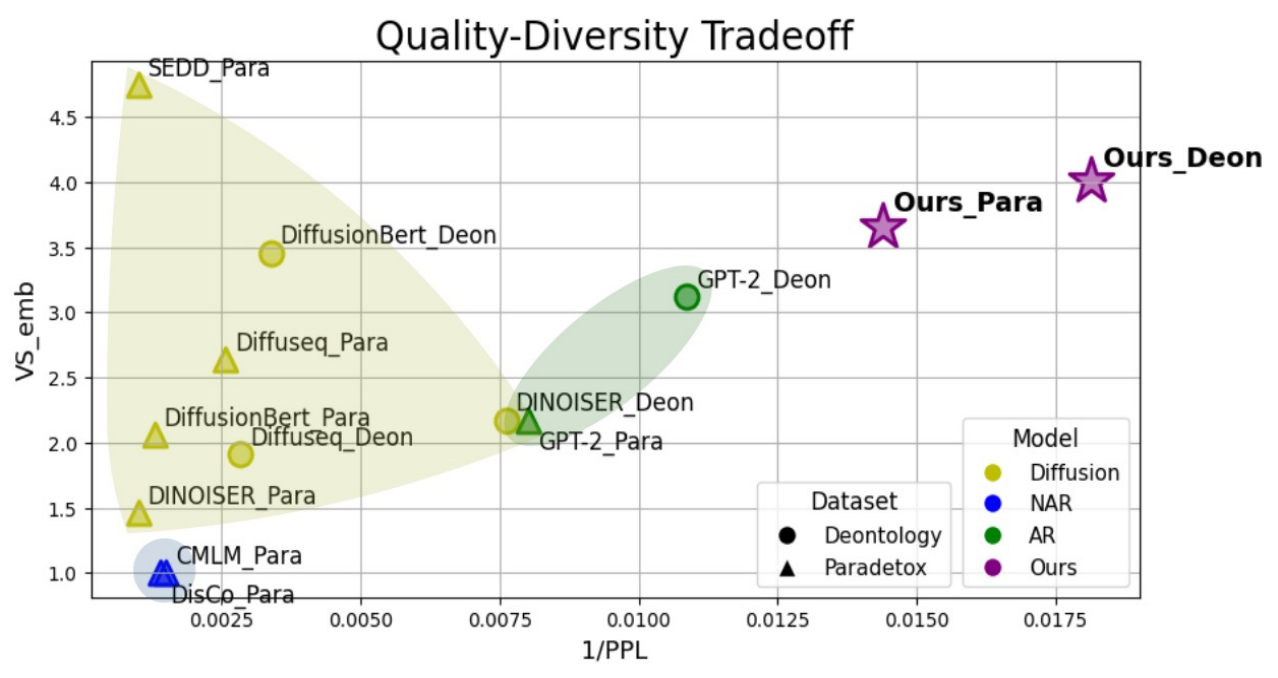}
\vspace*{-0.3cm}
\vspace{-0.5cm}
\caption{Quality--diversity tradeoff across various models. The x-axis ($1/\text{PPL}$) reflects generation quality, while the y-axis (VS\textsubscript{emb}) indicates diversity. Green points represent AR models, yellow points represent diffusion models, and blue points represent CMLMs. Our Diffusion-EAGS variants, marked by purple stars, achieve the best overall tradeoff.}
\vspace*{-0.6cm}
\label{fig:sec4}
\end{figure}

Figure~\ref{fig:sec4} presents the quality-diversity tradeoff among various models, where \textit{quality} is measured using perplexity (PPL) on the x-axis (inverted as $1/$PPL for better visualization) and \textit{diversity} is quantified using VS\_emb on the y-axis. Our model (\textbf{Ours\_Deon, Ours\_Para}, marked with purple stars) achieves the best tradeoff, outperforming prior approaches in both high-quality generation and diversity. 
Compared to Diffuseq, DiffusionBERT and CMLMs, our method achieves significantly better diversity without compromising generation fluency. This improvement stems from our \textbf{Entropy-Adaptive Gibbs Sampling (EAGS)}, which ensures controlled token selection, and \textbf{Entropy-based Noise Scheduling (ENS)}, which stabilizes the generation process. The results highlight that integrating MLMs into the diffusion framework enables high-quality, diverse, and controllable text generation.

\subsection{Keyword Based Generation}
\begin{table*}[h]
\centering
\tiny
\resizebox{\linewidth}{!}{%
\renewcommand{\arraystretch}{1.2}
\begin{tabular}{l ll}
\specialrule{0.5pt}{0pt}{0pt} \hline
                         % \multicolumn{3}{c}{Sentence}                                                   \\ \hline
\multicolumn{2}{l}{\multirow{2}{*}{\textit{Context}}}                       & \textit{Jake was playing with his toys. He accidentally broke his favorite one.}         \\
\multicolumn{2}{l}{}                                               & \textit{He cried a lot over it. His parents decided to replace it for him.}              \\ \hline
\multirow{3}{*}{Keyword } & \multicolumn{1}{l}{\textbf{not stop}}    & Jake just could \textbf{not stop} crying.                                               \\
                         &  \multicolumn{1}{l}{\textbf{Jake feel}}   & It made \textbf{Jake feel} So much better.                                              \\
                         & \multicolumn{1}{l}{\textbf{would enjoy}} & Jake said he \textbf{would enjoy} the new toy                                            \\ \hline\hline
\multicolumn{2}{l}{\multirow{2}{*}{\textit{Context}}}                       & \textit{Neil was in Sofia, Bulgaria. He was enjoying a trip backpacking through Europe.} \\
\multicolumn{2}{l}{}                                               & \textit{... He thought the food and culture in Sofia were the best.}                     \\ \hline
Keyword                  & \multicolumn{1}{l}{\textbf{Bulgaria!}}            & Things were looking great in \textbf{Bulgaria!}                                          \\ \hline\hline
\multicolumn{2}{l}{\multirow{2}{*}{\textit{Context}}}                       & \textit{Karen wanted to go on a trip to France. She started doing research on the trip.} \\
\multicolumn{2}{l}{}                                               & \textit{She decided to book a week long trip. She left the next day for her trip}sx.       \\ \hline
Keyword                  & \multicolumn{1}{l}{\textbf{her trip}}             & She spent almost a week there during \textbf{her trip}.                                  \\ \specialrule{0.5pt}{0pt}{0pt} \hline
\end{tabular}
}

\vspace*{-0.2cm}
\caption{Examples of keyword-based generation. Each row shows a \textit{Context} and a specified \textit{Keyword}, which is inserted into a masked position. The resulting outputs demonstrate how our model seamlessly integrates keywords into coherent narratives.}
\label{tab:keyword_gen_full}
\vspace*{-0.6cm}
\end{table*}
 Our model operating within discrete space enables us to manipulate the output sequences using explicit instructions. To further explore this capability, we conduct the generation of sequences based on keywords positioned in the middle and at the end of masked sequences, which is challenging for AR models~\citep{keskar2019ctrlconditionaltransformerlanguage}. They inherently struggle with controllability due to their inability to revise past steps based on future ones—an inductive bias of AR models. Initially, we provide the same contextual input while varying the keywords. In the masked states, we randomly select positions, replacing them with the specified keywords. The results in Table~\ref{tab:keyword_gen_full} demonstrate that the generated sequences seamlessly integrate the keywords with context-specific semantics. 
 
\subsection{Ablation Study}

\begin{table}[h]
\centering
% \tiny
\resizebox{\linewidth}{!}{%
\renewcommand{\arraystretch}{1.2}
\begin{tabular}{ll ccc cc}
\specialrule{1pt}{0pt}{0pt} \hline \toprule

                & Dataset   & PPL           & MAUVE         & SOME         & VS(ngram)     & VS(emb)    \\ \midrule
\multirow{2}{*}{Diffusion-EAGS} & Deont    & 55.1 & 0.412 & 0.835 & 4.898 & 4.009          \\
\multicolumn{1}{c}{}      & Roc      & 67.3     & 0.87   & 0.844                             & 4.837            & 3.999        \\ \midrule
\multirow{2}{*}{\textbf{w/o EAGS}} & Deont      & 667.9       & 0.022         & 0.617        & 4.767         & 3.928 \\
\multicolumn{1}{c}{}      & Roc      & 1084.9       & 0.035         & 0.613        & 4.874         & 3.957       \\  \midrule
\multirow{2}{*}{\textbf{w/o Gibbs Sampling}} & Deont      & 1426.7         & 0.011        & 0.584        & 2.378         & 1.923 \\
\multicolumn{1}{c}{}      & Roc      & 1293.1         & 0.010        & 0.534        & 1.531         & 1.338        \\  \midrule
\multirow{2}{*}{\textbf{w/o Pre-trained MLM}} & Deont      & $\geq$2K      & 0.005       & 0.645        & 4.758         & 3.402      \\
\multicolumn{1}{c}{}      & Roc      & $\geq$2K      & 0.004       & 0.604        & 4.315         & 2.994     \\ 
\specialrule{1pt}{0pt}{0pt} \hline
\end{tabular}%
}
\vspace*{-0.2cm}
\caption{Ablation study on the Deontology (Deont) and RocStories (Roc) datasets. ``w/o EAGS'' uses naive Gibbs sampling (no entropy estimation), ``w/o Gibbs Sampling'' removes diffusion process, and ``w/o Pre-trained MLM'' omits the pre-trained MLM entirely.}
\vspace{-0.3cm}
\label{tab:ablation}
\end{table}

To explore the effectiveness of our model's components, we conduct ablation studies focusing on three key elements: EAGS, Gibbs Sampling, and pre-trained MLM in Table~\ref{tab:ablation}.

% EAGS - Gibbs sampling - MLM 순으로 배치

The result of w/o EAGS shows a severe decline in text quality, \textit{producing degenerated results similar to those of traditional CMLMs}. Such phenomenon suggests that the naive application of MLM within the diffusion process fails to fully harness the capabilities of it.

Next, removing the use of the diffusion generation process (w/o Gibbs Sampling) leads to a drastic reduction in overall performance, with increased PPL and reduced diversity scores. These results imply that relying solely on MLM for text generation introduces considerable limitations.
% poses significant challenges.

Without the pre-trained MLM, outputs become highly degenerated, underscoring the need for precise entropy estimation.

% Before diffusion-EAGS, we consider three potential strategies for selecting the denoising target position based on entropy information (entropy estimation): 
In the process of selecting our highest-entropy-based scheduling in Diffusion-EAGS, we considered three alternatives: lowest entropy selection, random position selection following ENS training, and highest entropy selection. 
% We conduct an additional experiment on the paradetox dataset, obtaining 
Experiment on the paradetox dataset yielded PPL scores of 1193, 183, and 112, respectively. A subsequent heuristic evaluation confirms that the quality aligns with these PPL values. Consequently, we adopt the highest-entropy-based selection strategy. The process of schedule selection is detailed in Appendix~\ref{appx:entropy_selection_criteria}.

With EAGS, our model shows a substantial performance improvement. To verify the effectiveness of our model in guiding stable energy reduction, we examine the entropy flow during the generation process in Appendix~\ref{appendix:entropy_flow}. Our findings demonstrate that EAGS contributes significantly to a gradual decrease in entropy, enabling the generation of sentences in a stable manner.

\vspace{-0.1cm}
\section{Conclusions \& Discussions}
In this work, we introduce Diffusion-EAGS, an approach that integrates MLMs with diffusion models for conditional generation, yielding improved text quality, enhanced diversity, broad applicability, and precise token-level control.

% \textbf{1. How about other PLMs ?}
% We conducted a toy experiment using T5 within the Paradetox. The results, however, showed no significant difference compared to GPT-2 fine-tuning performance in Appendix~\ref{appdx:fg_comparison} Table~\ref{tab:t5_performance}. We conjecture that T5's generation process is already largely influenced by its initial decoder tokens from an entropy perspective~\cite{wang2024chainofthoughtreasoningprompting}, resulting in relatively low diversity. This finding implies that establishing a new theoretical connection between T5 and the diffusion process represents a another direction for future research on conditional generation tasks to broaden the applicable area of diffusion models.
\paragraph{Investigation of Other PLMs}  
% We conducted a toy experiment using T5 within Paradetox, but the results showed no significant improvement over GPT-2 fine-tuning (Appendix~\ref{appdx:fg_comparison}, Table~\ref{tab:t5_performance}). We hypothesize that T5's generation is highly influenced by its initial decoder tokens~\cite{wang2024chainofthoughtreasoningprompting}, leading to lower diversity. 
We conducted a toy experiment using T5 on the Paradetox dataset; however, the results showed no significant improvement over GPT-2 fine-tuning (see Appendix~\ref{appdx:fg_comparison}, Table~\ref{tab:t5_performance}). We hypothesize that T5's generation is heavily influenced by its initial decoder tokens~\cite{wang2024chainofthoughtreasoningprompting}, which leads to lower diversity.
% This suggests that establishing a theoretical connection between T5 and diffusion models could be an interesting direction for future research in conditional generation. By devising methodologies that align the training objectives of other PLMs with diffusion loss—similar to our approach—we can further accelerate advancements in diffusion-based NLP.
This suggests that developing a theoretical framework to integrate encoder-decoder models with diffusion processes may be a promising direction for future research in conditional generation. By devising methodologies that align the training objectives of other PLMs with diffusion loss—similar to our approach—, we can further accelerate progress in diffusion-based NLP.
% \textbf{2. Diversity for LLMs}
% The diversity of samplings has been shown to improve reasoning performance~\cite{lee2025generatingdiversehypothesesinductive}. However, increasing diversity solely through LLMs has inherent limitations in such research, which is also demonstrated by our experimental results in Appendix~\ref{appdx:diversity_saturation}. These findings indicate that diffusion models may serve as promising candidates for assisting LLMs, particularly in boosting reasoning performance through increased diversity.

% \vspace*{-0.2cm}
% \section{Conclusions}
% \vspace*{-0.2cm}
% \input{09_conclusion}

\section*{Limitations}

While Diffusion-EAGS demonstrates significant improvements in conditional generation tasks, there are several limitations. First, as our method is currently focused on text generation tasks, its applicability to text classification tasks, such as Named Entity Recognition and Part-of-Speech Tagging, remains unexplored. Future research could explore extending this method to other NLP tasks. Second, although our current efforts concentrate on developing and validating our framework using MLM, the potential integration of ARMs remains unexplored. With a proper methodology that aligns AR pre-training and diffusion training objectives, AR models would be another good initialization. Third, although the bias embedded in pre-trained models can be directly propagated, recent studies show that data-balancing strategies can effectively address this issue. Consequently, it is essential to account for these factors when deploying such models.

% \section*{Ethical Statements}
% \input{90_ethics}
% \section*{Acknowledgements}

% This document has been adapted
% by Steven Bethard, Ryan Cotterell and Rui Yan
% from the instructions for earlier ACL and NAACL proceedings, including those for 
% ACL 2019 by Douwe Kiela and Ivan Vuli\'{c},
% NAACL 2019 by Stephanie Lukin and Alla Roskovskaya, 
% ACL 2018 by Shay Cohen, Kevin Gimpel, and Wei Lu, 
% NAACL 2018 by Margaret Mitchell and Stephanie Lukin,
% Bib\TeX{} suggestions for (NA)ACL 2017/2018 from Jason Eisner,
% ACL 2017 by Dan Gildea and Min-Yen Kan, 
% NAACL 2017 by Margaret Mitchell, 
% ACL 2012 by Maggie Li and Michael White, 
% ACL 2010 by Jing-Shin Chang and Philipp Koehn, 
% ACL 2008 by Johanna D. Moore, Simone Teufel, James Allan, and Sadaoki Furui, 
% ACL 2005 by Hwee Tou Ng and Kemal Oflazer, 
% ACL 2002 by Eugene Charniak and Dekang Lin, 
% and earlier ACL and EACL formats written by several people, including
% John Chen, Henry S. Thompson and Donald Walker.
% Additional elements were taken from the formatting instructions of the \emph{International Joint Conference on Artificial Intelligence} and the \emph{Conference on Computer Vision and Pattern Recognition}.

% Bibliography entries for the entire Anthology, followed by custom entries
%\bibliography{anthology,custom}
% Custom bibliography entries only
% \section*{Acknowledgements}
% % \input{95_acknowledgements}
\bibliography{custom}

% Ref

%\bibliography{anthology,ref}

% Appendix
\clearpage
\newpage
\appendix
\maketitle

\section{Necessity of Energy Update in cMRF Generation}
\label{appx:need_engergy}
We observe a significant increase in log-potential values for sequences when guided by the RocStories conditions, as shown in Figure~\ref{fig:potential}. Additional experiments supporting this observation are detailed in Appendix~\ref{appx:potential function}.
\vspace{-0.1cm}
\begin{figure}[h]
    \centering
    \vspace{-0.1cm}
    \includegraphics[width=\linewidth]{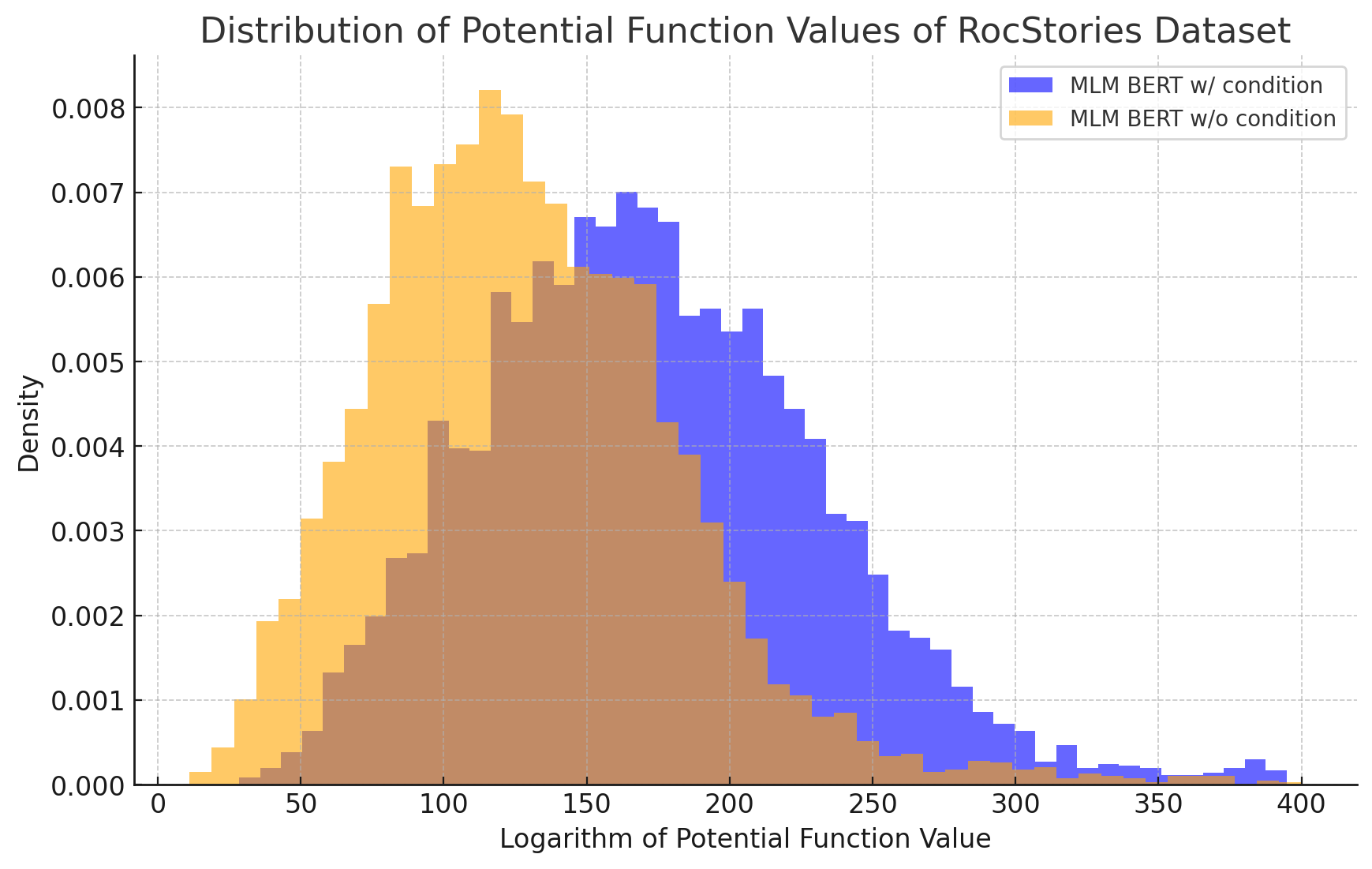}\vspace*{-0.3cm}
    \caption{When a condition is provided, the distribution of potential values for the samples is shifted on a logarithmic scale.}
    \label{fig:potential}
    
\end{figure}

This observation implies that conditional sequences differ from different conditional sequences in terms of randomness, 
making it crucial to update the energy function when the conditioning changes. 
For instance, \emph{MASK MASK author} and \emph{MASK am author} belong to different random fields, 
as also suggested by \citet{goyal2022exposingimplicitenergynetworks}.

\section{Measuring Potential Function in MLM}
\label{appx:potential function}

In this section, we provide additional experimental details and results to support the observation that open-ended Masked Language Models (MLMs) exhibit increased potentials for the same sequence under different dataset constraints.

\subsection{Experimental Setup}

\begin{itemize}
    \item \textbf{Model} We use the pre-trained BERT large model (\texttt{bert-large-cased}) as the base model for all experiments. Additionally, we incorporate RocStories-conditioned guidance with the pre-trained model and use a fine-tuned BERT model on the RocStories dataset to evaluate the impact of dataset-specific constraints.    
    \item \textbf{Tokenization} Tokenization is performed using the BERT tokenizer with special tokens (\texttt{[CLS]} and \texttt{[SEP]}).
    \item \textbf{Potential Calculation} The the log-potentials are obtained for each token using masked token logits.
    \item \textbf{Datasets}
          \begin{itemize}
              \item \textbf{RocStories:} Structured narratives from the RocStories dataset.
          \end{itemize}
\end{itemize}

\subsection{Results of Experiment and Implications for Conditional Generation}
Using the BERT-large-cased model, the average log potential value for the standard MLM was 156.6150, while incorporating RocStories guidance increased this value to 175.5332, highlighting the impact of dataset-specific constraints. Additionally, fine-tuning the same model on RocStories resulted in an average potential function value of 3.7551 (on an exponential scale), demonstrating substantial variation introduced by conditional generation settings. 

 The results demonstrate the significant influence of dataset structure on the potential function in MLMs. Specifically, structured datasets like RocStories enforce stronger narrative constraints, leading to higher potentials and greater coherence in sequence generation.

\section{The Candidates of Denoising Schedules}
\label{appx:entropy_selection_criteria}
We arrived at our proposed approach by going through several steps. The core of DDLM lies in how to define the denoising matrix.

\paragraph{1. Initial BERT Refinement Without a Noise Matrix}
We first explored a BERT-refinement method without a noise matrix, applying the same procedure at every step. Unsurprisingly, we found that the model failed to denoise the [MASK] tokens, resulting in sequences such as:

\begin{small}
\begin{verbatim}
[MASK] [MASK] educated ... educated [MASK] [MASK]
\end{verbatim}
\end{small}

\paragraph{2. BERT Denoising Matrix (0.15 Masking Ratio)}
 
Next, we implemented the denoising matrix using a BERT Denoising Matrix (0.15 Masking Ratio, \(1 - \frac{1}{T})\), which led to a strong bias toward a single repeated token:

\begin{small}
\begin{verbatim}
wwii wwii wwii wwii wwii wwii wwii wwii
\end{verbatim}
\end{small}

\paragraph{3. Time-Reversal Denoising (Tweedie-Leaping)}
Inspired by prior literature \cite{lou2024discrete}, we then examined a Time-Reversal Denoising Schedule Tweedie \(\tau\)-leaping based on score entropy. However, in the paradetox SEDD experiments, we observed NA results under strict conditional generation settings.

\paragraph{4. Word-Frequency-Based Denoising Schedule}
Subsequently, we applied a word-frequency-based denoising schedule \cite{he2022diffusionbert}, but in the paradetox DiffusionBERT experiments, this approach encountered difficulties in constructing coherent sentences.

\paragraph{5. Vocab-Wise Entropy Estimation}
\label{appdx:entropy_guide}
Moving on, instead of relying on word frequency, we propose a vocab-wise entropy estimation technique. In particular, we construct the denoising matrix as shown in 2, leveraging entropy information to decide whether each word should be denoised or preserved. This approach assumes that all positions, including originally masked ones, can potentially be denoised. Although this approach did show some improvement, for instance producing:

\begin{small}
\begin{verbatim}
wwii reassure wwii bony wwii wwii wwii wwii
\end{verbatim}
\end{small}

Upon further analysis, we identified that the MLM was not effectively determining which positions to denoise, and well-generated tokens sometimes are converted [MASK], and then convert all [MASK] tokens into certain words in the final step, leading to token replication.

\paragraph{6. Entropy-Based Estimation and Denoising}
Hence, we introduced an entropy-based estimation and denoising strategy. In this approach, we assume that once a mask is denoised, it remains fixed. Specifically, we select mask positions based on an entropy schedule, sample tokens for those positions, and once a token is sampled (i.e., denoised), we preserve it across subsequent diffusion steps. 

\paragraph{7. Entropy Selection Criteria}
We conducted three main experiments—uniform, reverse-order-EAGS, and EAGS—yielding perplexities of 182.976 with some portion of [MASK], 1193.229 with degenerated results, and 112.190 for paradetox dataset, respectively. These results indicate that noising from the most determinative token positions (mask with lowest entropy) is highly effective. Therefore, we adopt the Selection Criteria as EAGS.

\begin{figure}[h]
    \vspace{-0.2cm}
    \centering
    \includegraphics[width=0.9\linewidth]{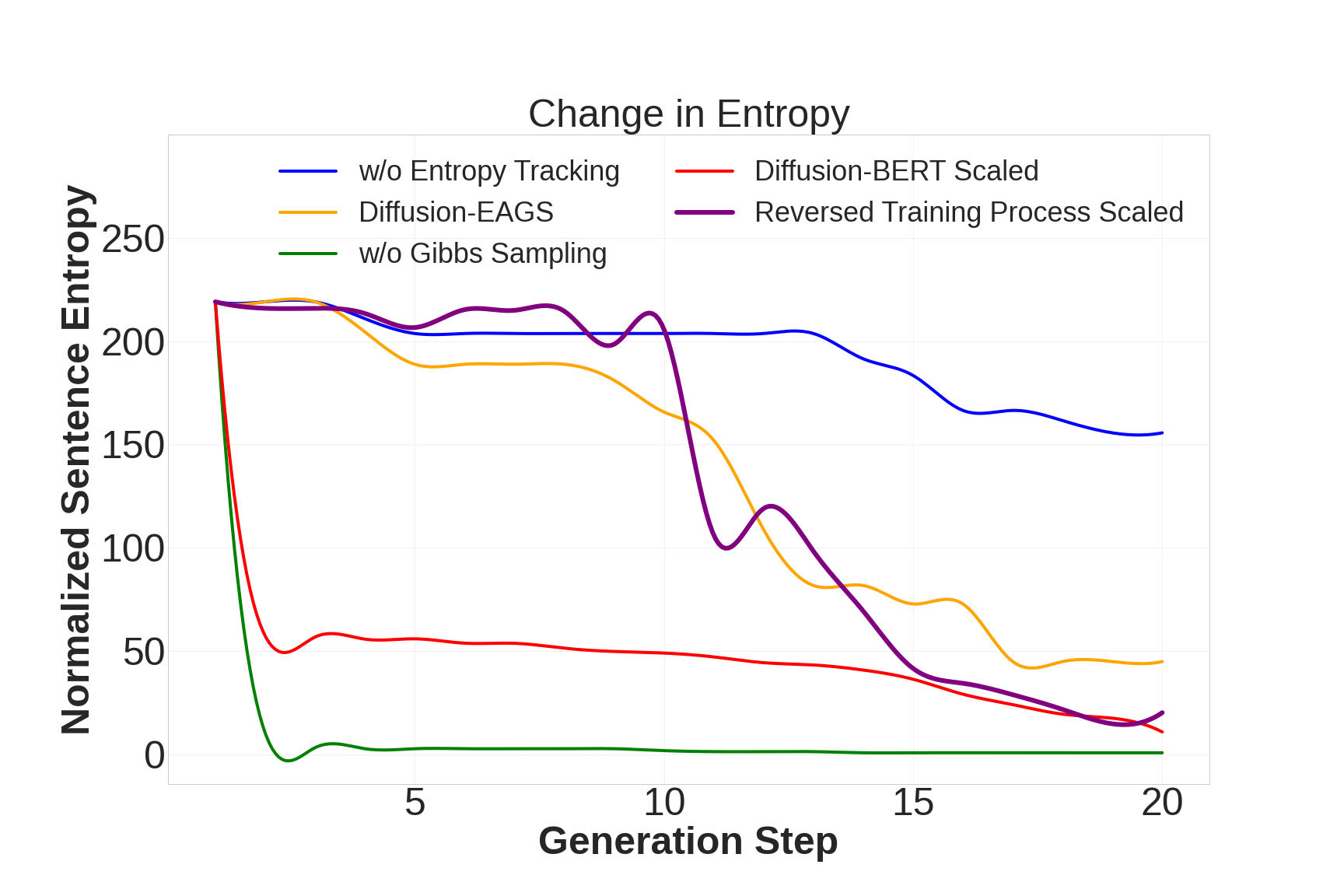}
    \vspace{-0.5cm}
    \caption{Entropy behavior tracking in generation/training process.}
    \label{fig:entropy}
    \vspace{-0.6cm}
\end{figure}

\section{Entropy Flow}
\label{appendix:entropy_flow}

In Figure~\ref{fig:entropy}, we illustrate the tendency of the sequential sum of entropy for various discrete generation processes. The changes of entropy during the generation process in Diffusion-EAGS, represented by the yellow line, show that our model effectively follows a gradual decrease in entropy, mirroring the inverse trend of the training process. This gradual change in entropy facilitates successful DDLM training, which results in superior text quality performance compared to other diffusion models, as demonstrated in Tables~\ref{tab:roc},~\ref{tab:social}, and~\ref{tab:qg_qqp}.

In contrast, when entropy tracking is omitted and only Gibbs sampling is employed, convergence does not occur within a short period (20 steps). The randomness of the sampling process leads to instability, resulting in lower average text quality, as shown in Table~\ref{tab:ablation}. Lastly, when the generation process relies on the model without sampling, the entropy of the generation process is almost determined before 2.5 steps. 
This entropy behavior is similar to that observed in DiffusionBERT.

\begin{algorithm}[h]
\footnotesize
\DontPrintSemicolon
\SetAlgoLined
\SetNoFillComment

\caption{EAGS Algorithm}
\label{alg:EAGS}
\textbf{EAGS Process:}\;
\textbf{Input:} Sequence Length $L$, Total Timestep $T$, Trained Model $M$, Mask Sequence Generator $G_M$, and Context $Y$\;

\For{$t = T$ to $0$}{
    \uIf{$t = T$}{
        $x^T \gets G_M(L, Y)$ \text{\scriptsize\color{blue}\itshape // Initialize a sequence of $L$}\;
    }
    \Else{
        $f_\theta \gets p_{\theta}(x^t, Y)$ \text{\scriptsize\color{blue}\itshape // Compute logits at timestep $t$}\;
        $l^* \gets \underset{l}{\arg\max}\; H(x^t_l \mid Y, f_\theta)$ \text{\scriptsize\color{blue}\itshape // Obtain nth largest entropy tokens ($M_t$)}\;
        $x^{t-1} \gets p_{\theta}(x^t, l^*, Y)$ \text{\scriptsize\color{blue}\itshape // Sample from the previous timestep}\;
    }
}
\end{algorithm}

\begin{algorithm}[h]
\footnotesize
\DontPrintSemicolon
\SetAlgoLined
\SetNoFillComment

\caption{ENS Algorithm}
\label{alg:ENS}
\textbf{ENS Process:}\;
\textbf{Input:} Context $Y$, Total Timestep $T$, and Dataset $D$\;

\For{$\text{Batch Step} = 0$ to $N$}{
    $x \sim D$ \text{\scriptsize\color{blue}\itshape // Sample data from \(D\)}\;
    $t \sim \text{Randint}(0, T)$ \text{\scriptsize\color{blue}\itshape // Sample random timestep}\;
    $f \gets \text{PLM}(x \mid Y)$ \text{\scriptsize\color{blue}\itshape // Compute logits using the PLM}\;
    $\mathcal{H} \gets H(x \mid Y, f)$ \text{\scriptsize\color{blue}\itshape // Calculate Entropy}\;
    $x^t \gets \text{Forward}(x_0, \mathcal{H}, t)$ \text{\scriptsize\color{blue}\itshape // Forward at $t$}\;
    $x^{t+1} \gets \text{Forward}(x_0, \mathcal{H}, t+1)$ \text{\scriptsize\color{blue}\itshape // Forward at $t+1$}\;
    $L_s = -\sum_{i} q(x^t_i \mid x_{t+1}) \log p_{\theta}(x^t_i \mid x_{t+1})$ \text{\scriptsize\color{blue}\itshape // Cross entropy loss calculation}\;
}
\end{algorithm}

\section{EAGS \& ENS algorithms}
Detailed algorithms of EAGS and ENS are in Algorithm~\ref{alg:EAGS} and~\ref{alg:ENS}.

\section{Experiment}
\label{appendix:exp}
\subsection{Fine-Grained Conditional Generation}
\label{appendix:datasets}
% \begin{table*}[t]
% \centering
% \tiny
% \vspace*{-0.3cm}
% \resizebox{\linewidth}{!}{%
% \begin{tabular}{l|cc|ccc}
% \toprule
% \textbf{Dataset/Benchmark} & \textbf{Open-ended Generation?} & \textbf{Conditional Generation?} & \textbf{Context provided?} & \textbf{Content provided?} & \textbf{Format provided?} \\
% \midrule
% Wikitext & $\checkmark$ & $\times$ & - & - & -  \\
% RocStories & $\checkmark$ & $\checkmark$ & \checkmark & \times & -  \\
% Deontology & $\triangle$ & $\checkmark$ & \checkmark & \triangle & \times  \\
% Question Generation & $\checkmark$ & $\checkmark$ & \checkmark & \checkmark & \times  \\
% Paraphrase & $\times$ & $\checkmark$ & \checkmark & \checkmark & \times  \\
% Translation & $\times$ & $\checkmark$ & \checkmark & \checkmark & \checkmark  \\
% ParaDetox & $\times$ & $\checkmark$ & \checkmark & \checkmark & \checkmark  \\
% \bottomrule
% \end{tabular}}
% \caption{Evaluation of NLP tasks in conditional-aspect: The first two columns assess general characteristics, while the last three columns measure the structuredness of the task (Context, Content, and Format provided). Tasks become more structured further down the table.}
% \vspace*{-0.4cm}
% \label{tab:data_cat}
% \end{table*}

\begin{table*}[t]
\centering
\tiny
\resizebox{\linewidth}{!}{%
\renewcommand{\arraystretch}{1.2}
\begin{tabular}{|l|c|c|c|c|c|c|c|}
\hline
\diagbox[width=2.8cm]{\makecell{\textbf{Type}}}{\makecell{\textbf{Dataset}}}
& \makecell{\textbf{RocStories}}
& \makecell{\textbf{Deontology}}
& \makecell{\textbf{Question} \\ \textbf{Generation}}
& \makecell{\textbf{QQP}}
& \makecell{\textbf{DialogSum}}
& \makecell{\textbf{ALMA}}
& \makecell{\textbf{ParaDetox}} \\ 
\hline
\textbf{Open-ended Generation}  & $\checkmark$ & $\triangle$ & $\checkmark$ & $\times$ & $\times$ & $\times$ & $\times$  \\
\textbf{Conditional Generation} & $\checkmark$ & $\checkmark$ & $\checkmark$ & $\checkmark$ & $\checkmark$ & $\checkmark$ & $\checkmark$  \\
\hline
\textbf{-- Context Provided ?}  & $\checkmark$ & $\checkmark$ & $\checkmark$ & $\checkmark$ & $\checkmark$  & $\checkmark$ & $\checkmark$  \\
\textbf{-- Content Provided ?}  & $\times$ & $\triangle$ & $\checkmark$ & $\checkmark$ & $\checkmark$ & $\checkmark$ & $\checkmark$  \\
\textbf{-- Format Provided ?} & - & $\times$ & $\times$ & $\times$ & $\triangle$ & $\checkmark$ & $\checkmark$   \\
\hline
\end{tabular}}
\caption{Each dataset has a different level of conditional constraints even if they are all conditional generation tasks. 
$\checkmark$ indicates full support, $\times$ indicates no support, and $\triangle$ indicates partial or limited support.}
\label{tab:data_cat}
\vspace*{-0.4cm}
\end{table*}

% \begin{table*}[t]
% \centering
% \tiny
% \resizebox{\linewidth}{!}{%
% \renewcommand{\arraystretch}{1.2}
% \begin{tabular}{|l|c|c|c|c|c|c|c|}
% \hline
% \diagbox[width=2.8cm]{\makecell{\textbf{Type}}}{\makecell{\textbf{Dataset}}} 
% & \makecell{\textbf{RocStories} \\ 
% & \makecell{\textbf{Deontology} \\ 
% & \makecell{\textbf{Question Generation} \\
% & \makecell{\textbf{QQP} \\ 
% & \makecell{\textbf{ALMA} \\ 
% & \makecell{\textbf{ParaDetox} \\ 
% \hline 
% \textbf{Open-ended Generation}  & $\checkmark$ & $\triangle$ & $\checkmark$ & $\times$ & $\times$ & $\times$  \\
% \textbf{Conditional Generation}  & $\checkmark$ & $\checkmark$ & $\checkmark$ & $\checkmark$ & $\checkmark$ & $\checkmark$  \\ \hline 
% \textbf{-- Context Provided ?}  & $\checkmark$ & $\checkmark$ & $\checkmark$ & $\checkmark$ & $\checkmark$ & $\checkmark$  \\
% \textbf{-- Content Provided ?}  & $\times$ & $\triangle$ & $\checkmark$ & $\checkmark$ & $\checkmark$ & $\checkmark$  \\
% \textbf{-- Format Provided ?} & - & $\times$ & $\times$ & $\times$ & $\checkmark$ & $\checkmark$  \\
% \hline
% \end{tabular}}
% \caption{Each dataset has a different level of conditional contraints even if they are conditional generation tasks. $\checkmark$ indicates full support, $\times$ indicates no support, and $\triangle$ indicates partial or limited support.}
% \label{tab:data_cat}
% \end{table*}

In conditional generation tasks, the level of conditional constraint imposed by the dataset plays a critical role in shaping the generation process. As shown in Table~\ref{tab:data_cat}, conditional constraints are diverse across datasets. In our task, we categorize these constraints into three levels: (1) the provision of context alone, requiring the continuity of the prefix; (2) the provision of specific content to be included in the target sequence, necessitating the inclusion of certain keywords; and (3) the provision of semantic content formatting, such as transforming toxic sentences into safer alternatives or converting text from the source language to a target language. In our study, we aim to develop a diffusion framework capable of being applied across a wide range of conditional generation tasks.

\subsection{Dataset Explanations}

\begin{table}[h]
\centering
\resizebox{\linewidth}{!}{%
\renewcommand{\arraystretch}{1.2}
\begin{tabular}{l|cc|cc|cc|cc|cc}  
\specialrule{1pt}{0pt}{0pt} \hline
 & \multicolumn{2}{c|}{Quasar-T} & \multicolumn{2}{c|}{QQP} & \multicolumn{2}{c|}{ParaDetox} & \multicolumn{2}{c|}{Deontology} & \multicolumn{2}{c}{RocStories} \\ 
\hline
 & input & output & input & output & input & output & input & output & input & output \\ 
\hline
Max & 63 &  244  & 104 & 98 & 35 & 35 & 24 & 31 & 76 & 57 \\ 
Mean & 14.574 & 31.157 & 13.947 & 13.956 & 15.135 & 13.035 & 13.039 & 12.548 & 42.189 & 13.307 \\ 
\specialrule{1pt}{0pt}{0pt} \hline
\end{tabular}%
}
\caption{Dataset Statistics}
\label{tab:data}
\vspace{-0.6cm}
\end{table}

\textbf{Open-ended Generation} We employ the RocStories dataset \citep{mostafazadeh2016corpus} for open ended generation with narrative understanding tasks. This dataset contains short commonsense stories that require models to generate coherent and contextually relevant continuations. Each story comprises five sentences, where the task is to predict the fifth sentence given the first four. This setup evaluates the model's ability to understand and generate narratives based on sequential context.

\textbf{Deontology} The objective of Deontology \citep{hendrycks2023aligning} is to evaluate the capability of models to make ethical judgments from a deontological perspective. The dataset contains scenarios focusing on interpersonal dynamics and everyday occurrences. 

\textbf{Paraphrase} The objective of the Quora Question Pairs (QQP) \citep{wang2017bilateral} is to determine whether two questions are paraphrases of each other. We process the QQP dataset by treating one question as a paraphrase of another, a method commonly employed to assess the effectiveness of diffusion models.

 \textbf{QG} The objective of Question Generation (QG) is to generate valid and fluent questions based on a given passage and a specified answer. We employ the Quasar-T dataset, introduced by \citet{dhingra2017quasar} in 2017, which comprises a substantial number of document-question pairs. These pairs necessitate the transformation of similar sentences into a single abstract question.

\textbf{DialogueSum} In former experiments, it is hard to measure the performance with reference-based metrics as the limitation of traditional EM problems where conditional generation’s output space is wide. Therefore, to test our model’s capability, we experiment on dialogue summarization task~\cite{chen2021dialogsumreallifescenariodialogue} which makes an emphasis on containing some keywords or necessary information in the generated sequences. We use the experimental dataset and evaluation metric proposed in DiffusionCG~\cite{xiang2024diffusiondialogdiffusionmodeldiverse} with same experimental setting as former experiments.

 \textbf{Machine Translation} Labeled datasets used in conditional generation tasks are typically limited in size and sometimes multilingual. To further assess our model's performance in conditional generation, particularly in terms of language extension and resource scarcity, we conduct additional experiments on a translation task. We utilize the 18k \textit{en}$\leftrightarrow$\textit{de} human-curated dataset by \citet{xu2024a,pmlr-v235-xu24t}.

\textbf{Paradetox} The objective of the Paradetox \citep{logacheva-etal-2022-paradetox} is to delete the profanities in source sentence. It comprises of toxic and neutral utterances, curated from the Jigsaw, Reddit, and Twitter datasets.

\subsection{Experimental Details}
\label{appdx:metrics_hyperparams}
We employ roberta-base as MLM with learning rate 5e-4. The maximum lengths for QG, QQP, and Paradetox is set to 64, while for Deontology and DialogSum set to 48 and 292, respectively, based on data statistics. We test 20 conditions with 5 outputs in total 100, which is not used for training. The number of steps is configured to 5. We then perform a naive categorical sampling with a sample size of 20 and select final 5 samples based on PPL. We use 1 A100 GPU with the batch size as 256. 

For the case of ARMs, CMLMs, CDLMs, and DDLMs, we follow the official repositories to reproduce the results. Results are sampled multiple times with different seeds to evaluate the diversity. For hyperparameters, we follow the original repositories if the parameter is provided, except for modifying the number of samples as 5 and max\_length parameters according to data statistics. Note that unlike other benchmarks, we experiment with Diffuseq-v2~\citep{gong2023diffuseqv2} in translation task for a broader comparison with existing baselines. Moreover, experimental details of LLMs are in Appendix~\ref{appedix_gpt}, machine translation in Appendix~\ref{appendix:expouts}.

\paragraph{Quality metrics} To measure the quality of the generated texts, we use Perplexity based-on  GPT-2 Large and GPT-2 XL, SOME \citep{yoshimura-etal-2020-reference}, the grammar metric based on corpus, LLM-c \citep{lin-chen-2023-llm} to measure the plausibility of the narratives, LLM-t \citep{koh2024llmsrecognizetoxicitydefinitionbased} to measure toxicity, and MAUVE \cite{pillutla2021mauve}, measuring a reflectiveness of training dataset characteristics of generate outputs. MAUVE score of 1 indicates that the output perfectly matches the training dataset as a neural database. For Mean Opinion Score (MOS), we get 5 outputs from each condition. For a fair MOS comparison, if GPT-3.5-turbo refuses to provide an answer or if sentence completeness is compromised by condition consisting of “rtttt,” or extreme elliptical expressions, we exclude such relevant condition from our evaluation target. Subsequently, hired four integrated ph.d course work annotators in the university NLP research lab evaluate the generated text based on two criteria: (1) semantic reflectiveness of the condition, indicating how accurately the condition is represented in the text, and (2) sentence completeness, assessing overall grammatical and semantic coherence. Each criterion was rated on a scale from 0 to 1. Subsequently, these scores are normalized and averaged to obtain a final score ranging from 0 to 1. In our evaluation, Fleiss’ kappa~\cite{fleiss1971measuring} is exceeded 0.7 as assessing sentence quality is both intuitive and relatively non-controversial among the annotators.

\paragraph{Diversity Metrics} Traditional diversity metrics Self-BLEU \cite{zhu2018texygen} and distinct-n \cite{li2015diversity} are employed to evaluate the generated texts. We also adopt Vendi Score (VS)-SimCSE~\cite{friedman2023vendi}, an interpretable diversity metric, which quantifies the effective number of unique samples in a given set. Both the n-gram and embedding variations are utilized, where embedding VS is semantic diversity. For the diversity MOS evaluation, we adopt the same methodology used for the quality MOS but apply two distinct criteria: (1) the condition’s semantic reflectiveness, and (2) sentence diversity, capturing both semantic and structural variety beyond mere word deletion or rearrangement. Ideal score of diversity MOS is 5 which means different five sequences for one condition, and lowest score is 1 which means all identical sequences.

\begin{table*}[h]
\centering
\tiny
\renewcommand{\arraystretch}{1.2}
\vspace*{-0.3cm}
\resizebox{\linewidth}{!}{%
\begin{tabular}{ll ccc ccccc} \specialrule{0.5pt}{0pt}{0pt} \hline
                           & \multicolumn{8}{c}{\textbf{ParaDetox}}      \\ \hline
                           \multicolumn{2}{c}{} & \multicolumn{3}{c}{Text Quality} &\multicolumn{5}{c}{Diversity} \\

Model                      & Step & PPL $\downarrow$          & MAUVE $\uparrow$        & SOME $\uparrow$         & VS(ngram) $\uparrow$    & VS(emb) $\uparrow$      & self-bleu $\downarrow$          & distinct-1 $\uparrow$  & distinct-2 $\uparrow$ \\ \hline
GPT-2                      & 1    & 389.1       & 0.503         & 0.717         & 3.925         & 2.640         & 0.429               & 0.312       & 0.748                 \\ 
GPT-3.5 w/ 4-shot                         & 1     & 104.375       & 0.175         & \textbf{0.888}                  & 3.098         & 1.915         & 0.652                                                   & 0.390       & 0.835            \\
GPT-4 w/ 4-shot                         & 1     & 78.979        & 0.125         & 0.879                   & 3.214         & 1.906         & 0.592                                                   & 0.412       & \textbf{0.841}          \\  \hline
CMLM w/ Mask-Predict &  10& 669.9 & 0.0234 & 0.588 & \textcolor{gray}{1.000} & \textcolor{gray}{1.000} & \textcolor{gray}{1.000} & \textcolor{gray}{0.451} & \textcolor{gray}{0.633} \\
DisCo w/ Easy-First & 10& 716.1 & 0.0344 & 0.576 &\textcolor{gray}{1.000}&\textcolor{gray}{1.000}&\textcolor{gray}{1.000}&\textcolor{gray}{0.438}&\textcolor{gray}{0.583}\\
DiffusionBert              &  2000    & 775.9       & 0.737         & 0.716         & \textcolor{gray}{3.101}         & \textcolor{gray}{2.058}         & \textcolor{gray}{0.599}               & \textcolor{gray}{0.424}       & \textcolor{gray}{0.826}       \\ \hline
DiffuSeq                   &  2000    & $\geq1k$      & \textcolor{gray}{0.683}         & \textcolor{gray}{0.703}         & \textcolor{gray}{2.059}         & \textcolor{gray}{1.465}         & \textcolor{gray}{0.841}               & \textcolor{gray}{0.410}       & \textcolor{gray}{0.820}                \\
LD4LG	                   &  2000    & 579.9       & 0.556         & 0.762          & 1.914         & 1.425          & 0.845        & 0.419	              & 0.829 \\
DINOISER                   &  20      & 124.8       & 0.255         & 0.767         & 2.287         & 2.174         & 0.981               & 0.211       & 0.486                \\ 
 SEDD                 &  1024    & $\geq1k$       & \textcolor{gray}{NA}        & \textcolor{gray}{0.664}         & \textcolor{gray}{4.746}         & \textcolor{gray}{4.063}         & \textcolor{gray}{0.119}               & \textcolor{gray}{0.451}       & \textcolor{gray}{0.846}       \\ \hline
\textbf{Diffusion-EAGS}      &  5    & 109.3 & 0.811 & 0.760 & 4.417 & 3.311 & 0.256 & 0.407 & 0.810         \\\specialrule{0.5pt}{0pt}{0pt} \hline
                           & \multicolumn{8}{c}{\textbf{Deontology}}      \\ \hline
                                                               & Step & PPL $\downarrow$ & MAUVE $\uparrow$ & SOME $\uparrow$ & VS(ngram) $\uparrow$ & VS(emb) $\uparrow$ & self-bleu $\downarrow$ & distinct-1 $\uparrow$ & distinct-2 $\uparrow$\\  \hline
GPT-2                           & 1 & 92.0  & 0.131 & \textbf{0.860} & 3.665 & 3.126 & 0.425 & \textbf{0.474} & \textbf{0.874} \\ 

DiffuSeq                        & 2000 & 352.8  & 0.005 & 0.703 & 2.273 & 1.915 & 0.753 & 0.267 & 0.745   \\
DINOISER                        & 20 & 131.3 & 0.008 & 0.740 & 2.287 & 2.174 & 0.824 & 0.309 & 0.713   \\ 
DiffusionBert                   & 2000 & 295.5 & 0.306 & 0.787 & 4.258 & 3.458 & 0.229 & 0.445 & 0.849 \\ 
%SEDD-small                      & 1024 & 155.8       & \textbf{0.858}         & 0.826         & 4.661         & \textbf{4.063}         & 0.112              & 0.437       & 0.850       \\
\hline
\textbf{Diffusion-EAGS}         & 5 & \textbf{55.1} & \textbf{0.412} & 0.835 &\textbf{4.898} & \textbf{4.009} & \textbf{0.056} & 0.418 & 0.806  \\ \specialrule{0.5pt}{0pt}{0pt} \hline
\end{tabular}}
\vspace*{-0.3cm}
\caption{Social Generation -- Diversity values associated with higher perplexity (PPL) are displayed in gray, as increased perplexity typically indicates degenerate sequences. }
\vspace{-0.1cm}
\label{tab:social}
\end{table*}

\begin{table*}[h]
\centering
\renewcommand{\arraystretch}{1.2}
\vspace*{-0.3cm}
\resizebox{\linewidth}{!}{%
\begin{tabular}{ll ccc ccccc}
\specialrule{0.5pt}{0pt}{0pt} \hline
                           & \multicolumn{8}{c}{\textbf{QQP}}      \\ \hline
Model                      & Step & PPL $\downarrow$          & MAUVE $\uparrow$        & SOME $\uparrow$        & VS(ngram) $\uparrow$    & VS(emb) $\uparrow$      & self-bleu $\downarrow$ & distinct-1 $\uparrow$ & distinct-2 $\uparrow$ \\ \hline
GPT-2                      & 1     & 66.270       & 0.112         & 0.754        & 3.886         & 2.566         & 0.423     & 0.344      & 0.787                \\ 
DiffuSeq                   & 2000     & 124.247       & 0.00674       & 0.709        & 1.927         & 1.242         & 0.813     & 0.226      & 0.543                \\
DINOISER                   & 20     & 79.742        & 0.0042        & 0.821        & 1.421         & 1.126         & 0.935     & 0.264      & 0.542       \\ 
DiffusionBert             & 2000     & 500.959       & 0.0709        & 0.618        & \textbf{4.489}         & \textbf{2.836}         &\textbf{0.196}     & 0.321      & 0.761       \\  \hline
\textbf{Diffusion-EAGS}         &  5     & \textbf{48.106}        & \textbf{0.683}         & \textbf{0.824}        & 4.006         & 2.390         & 0.338     & \textbf{0.421}      & \textbf{0.832}       \\ 
\specialrule{0.0pt}{0pt}{0pt} \hline
                           & \multicolumn{8}{c}{\textbf{QG}}      \\ \hline

Model                      & Step  & PPL $\downarrow$          & MAUVE $\uparrow$          & SOME $\uparrow$               & VS(ngram) $\uparrow$     & VS(emb) $\uparrow$       & self-bleu $\downarrow$     & distinct-1 $\uparrow$    & distinct-2 $\uparrow$    \\  \hline
GPT-2                      & 1      & 124.8         & 0.141           & 0.759       & 4.564          & 3.130          & 0.176          & 0.210          & 0.629          \\ 
DiffuSeq                   & 20     & 395.0          & 0.149          & 0.730              & 1.555          & 1.274          & 0.901          & 0.170          & 0.564          \\
DINOISER                   & 2000   & 155.9   & \textbf{0.159}                 & 0.776               & 1.396          & 1.121          & 0.944          & 0.166          & 0.553          \\  
DiffusionBert             & 2000     & 513.6         & 0.150          & 0.712               & 3.040          & 2.209          & 0.566          & \textbf{0.392}          & 0.759          \\ \hline
\textbf{Diffusion-EAGS}       & 5    & \textbf{80.7}    &  0.121          & \textbf{0.782}    & \textbf{4.646} & \textbf{3.538} & \textbf{0.152} & \textbf{0.403}          & \textbf{0.798}          \\ 
\specialrule{0.5pt}{0pt}{0pt} \hline
\end{tabular}}
\vspace*{-0.3cm}
\caption{QG \& QQP Generation}
\vspace*{-0.4cm}
\label{tab:qg_qqp}
\end{table*}

\begin{table*}[h]
\centering
\small  % Make the font a bit smaller if needed
\begin{tabular}{lcccccccc}
\toprule
\textbf{Model} & \textbf{ROUGE-1} & \textbf{ROUGE-2} & \textbf{MAUVE} & \textbf{Ngram} & \textbf{Emb} & \textbf{Self-BLEU} & \textbf{Distinct-1} & \textbf{Distinct-2} \\
\midrule
Ours      & 0.409 & 0.174 & 0.536 & 4.114 & 2.591 & 0.252 & 0.253 & 0.632 \\
SEDD      & 0.179 & 0.032 & 0.999 & 4.216 & 2.576 & 0.211 & 0.200 & 0.609 \\
DINOISER  & 0.209 & 0.031 & 0.337 & 1.247 & 1.227 & 0.926 & 0.256 & 0.633 \\
\bottomrule
\end{tabular}
\caption{DialogueSum Experiment}
\label{tab:dialoguesum}
\vspace*{-0.4cm}
\end{table*}

\section{Detailed analysis of Results}
\label{appdx:results}

\subsection{Fine-Grained Comparison}
\label{appdx:fg_comparison}
As shown in Table~\ref{tab:roc},~\ref{tab:social},~\ref{tab:qg_qqp}, our model consistently exhibits exceptional performance in terms of text quality while simultaneously maintaining diversity when compared to baseline models. The standard deviation of PPL in Paradetox Experiment is 61 for our model. All other PPL's standard deviation are similar to that of Paradetox.

In Table~\ref{tab:social} Paradetox, our model demonstrates superior performance across all evaluated metrics. Such phenomenon represents that our model based on MLM shows robustness on diverse perturbations of daily dialogues. When PPL exceeds 600, the model is considered to have failed in generating natural sequences, and is thus represented in gray color. Specifically, the text quality produced by the CMLM, which is standard BERT-generation, and SEDD, which is powerful model in open-ended generation, is found to be low. 

Consequently, these models were excluded from subsequent experiments. In Deontology, our model exceeds the baseline models' PPL and MAUVE scores, whereas SOME score represent the sufficient quality of text with the highest diversity score. As illustrated in Table~\ref{tab:qg_qqp}, Diffusion-EAGS generates the responses with the highest PPL score for QG, and highest MAUVE and PPL score for QQP.

While we adhere to the standard metrics commonly used in diffusion research and integrate as many additional metrics as possible, we also comprehensively explore our model’s capabilities across multiple dimensions. As the outputs of earlier generation tasks are too broad to be effectively evaluated using reference-based metrics, we provide generated examples in Appendix~\ref{appdx:generated_examples} and measure the preference of these outputs using a LLM-based metric in Appendix~\ref{appdx:human_score_rouge_score}. Additionally, to accommodate a scenario where reference-based evaluation is applicable, we have included a more extensive summarization task in Appendix~\ref{appdx:human_score_rouge_score} and translation task in Appendix~\ref{appx:translation_fullexperiment}. These results confirm that our method consistently produces outputs that adhere to the specified conditions.

Diffusion-EAGS demonstrates the highest MAUVE score in Table~\ref{tab:social}-ParaDetox, and high level of text quality surpassing that of GPT-2 in Table~\ref{tab:qg_qqp} in text quality. ParaDetox is colloquial dataset including slang, numerous abbreviations, and various perturbations, so our model demonstrate robustness to such perturbations. As for diversity, our model consistently outperforms GPT models in VS(ngram) and VS(emb) in Table~\ref{tab:roc},~\ref{tab:social}, and~\ref{tab:qg_qqp}. 

Notably, CDLMs demonstrate a noticeable deficiency in diversity. Examining the results of Diffuseq, it is evident that the grammar score is comparatively lower than that of other models. This outcome is expected, as the outputs from Diffuseq frequently display inaccurate sentence structures, including duplications of words or phrases. Conversely, the outputs from Dinoiser achieve moderate grammar scores but show limited diversity. This finding, coupled with our additional experiments concerning the beam size during Dinoiser generation, suggests that Dinoiser's performance predominantly relies on memorization. In contrast, our model excels at producing significantly more diverse sequences. Furthermore, our models require only a few steps, while resulting in higher quality and diversity.

\subsection{Quality Recheck -- LLM score \& Dialogue Summarization}
\label{appdx:human_score_rouge_score}

\begin{table}[H]
\centering
\renewcommand{\arraystretch}{1.2}
\resizebox{\linewidth}{!}{%
\begin{tabular}{lccccccc}
\toprule
\textbf{Model} & \textbf{PPL} & \textbf{MAUVE} & \textbf{vs(ngram)} & \textbf{VS(emb)} & \textbf{sef-bleu} & \textbf{distinct-1} & \textbf{distinct-2}\\
\midrule
GENIE & 134.1 & 0.296 & 2.527 & 1.800 & 0.702 & 0.454 & 0.825\\
\bottomrule
\end{tabular}
}
\caption{Quantitative results for the GENIE model.}
\label{tab:genie-results}
\end{table}
\vspace{-0.4cm}
\begin{table}[H]
\centering
\begin{tabular}{l|c}
\specialrule{1pt}{0pt}{0pt} 
\hline
& \textbf{LLM-t} \\ 
\hline \hline
GPT-2            & 0.02       \\
GPT-3.5          & 0.074      \\
GPT-4            & 0.18       \\
DiffuSeq         & \textcolor{gray}{0.03}   \\
Diffusion-Bert   & 0.09       \\
DINOISER         & 0.1        \\
SEDD-small       & \textcolor{gray}{NA}     \\
\textbf{Diffusion-EAGS} & \textbf{0.01}   \\
\specialrule{1pt}{0pt}{0pt}
\end{tabular}

\caption{\textbf{ParaDetox Dataset Generation} -- LLM-t is the LLM-evaluation for measuring toxicity.}
\label{tab:paradetox_full}
\vspace{-0.4cm}
\end{table}

\paragraph{Paradetox w/ LLM-t on application models}

Since our research primarily aims to enhance the model's inherent capabilities, we set up baselines that revolve around (or are closely related to) noise scheduling. Nevertheless, some studies employ a hybrid framework integrating LLMs and diffusion models \cite{lin2023textgenerationdiffusionlanguage,xiang2024diffusiondialogdiffusionmodeldiverse}; 
hence, we conduct additional experiments to investigate this scenario. In addition, to evaluate the quality of the \textsc{paradetox} output and ours diffusion-EAGS still outperforms GENIE~\citep{lin2023textgenerationdiffusionlanguage} in Table~\ref{tab:genie-results}. We also use the LLM-t score~\cite{koh2024llms} to measure whether models successfully detoxify the source condition, showing the quality of generated outputs from ours as shown in Table~\ref{tab:paradetox_full}.

\begin{table}[H]
\centering
\scriptsize
\begin{tabular}{l ccc}
\toprule
\textbf{Models} & \textbf{Prefer Baseline} & \textbf{Prefer Ours} & \textbf{Tie} \\
\midrule
diffuseq vs.\ ours       & 20\% & 65\% & 15\% \\
diffusionBERT vs.\ ours  & 20\% & 65\% & 15\% \\
dinoiser vs.\ ours       &  0\% & 90\% & 10\% \\
GPT-2 vs.\ ours          & 25\% & 65\% & 10\% \\
\bottomrule
\end{tabular}
\caption{Evaluation results comparing our model with various baselines.}
\label{tab:evaluation-results}
\vspace{-0.5cm}
\end{table}
\paragraph{QG - LLM preference}
For Question Generation (QG), we employ the widely adopted GPT-as-a-Judge framework~\cite{zheng2023judging} to evaluate the quality of generations produced by our model and the baselines on the QG dataset. We adopt a pairwise evaluation setting, following the system and input prompts specified in \citet{zheng2023judging} for the pairwise comparison. The factors specified to be evaluated are 1) coherency 2) grammatical correctness 3) semantic soundness 4) diversity and 5) being a more reasonable question to the input (condition) text. We employ the gpt-4 model. The result is in Table~\ref{tab:evaluation-results}.

Note that since within the prompt, the baseline model’s generations are specified prior to our model’s generation, there is a significant position bias working against our favor, as noted in \citet{zheng2023judging}. The results above indicate that despite such bias, our model’s generations are much more favored over the baselines’ generations.

\paragraph{Dialoguesum Experiment} 
Our model outperforms existing baselines in ROUGE, a reference-based metric as shown in Table~\ref{tab:dialoguesum}. These findings indicate that, according to the automatic scores, our model sufficiently captures the source condition.

\paragraph{Human Evaluation}
Below, we report the Mean Opinion Score (MOS) averages and standard deviations (std) in the following order: DiffusionBERT, LD4LG, GPT-2, Dinoiser, and our method.
First, the average scores of semantic reflection are 0.98, 0.90, 0.94, 0.98, and 0.97, respectively, with standard deviations of 0.14, 0.30, 0.24, 0.14, and 0.16. Second, the average scores of sentence completeness are 0.78, 0.92, 0.72, 0.84, and 0.90, respectively, with standard deviations of 0.18, 0.14, 0.28, 0.15, and 0.15. Third, average scores of diversity are 2, 1, 2.65, 1, and 4.6, respectively, with standard deviations of 1.3, 0, 1.45, 0, and 0.7. GPT-3.5-turbo's std is 0 for quality MOS and 0.83 for diversity MOS.

\begin{table}[h]
\centering
\scriptsize

\resizebox{0.95\linewidth}{!}{%
\renewcommand{\arraystretch}{1.2}
    \begin{tabular}{llccc}
        \toprule
        \multicolumn{1}{c}{\textbf{Model}} & \textbf{SacreBLEU} & \textbf{COMET} & \textbf{XCOMET} \\
        \midrule
        DisCo & &  & \\
        \hspace{0.1cm} w/ Easy-First   & 3.2806 & 0.2447 & 0.2414 \\
        \hspace{0.1cm} w/ Mask-Predict & 3.2862 & 0.2444 & 0.2414 \\
        DisCo-m   &  &  & \\        
        \hspace{0.1cm} w/ Easy-First   & 3.7423 & 0.2468 & 0.2122 \\
        \hspace{0.1cm} w/ Mask-Predict & 3.7748 & 0.2466 & 0.2119 \\
        \midrule
        Diffuseq-v2 & 1.90 & 0.3242 & 0.2628 \\
        SEDD & & & \\
        \hspace{0.1cm} w/ from scratch & 0.14 & 0.2375 & 0.2035 \\
        \hspace{0.1cm} w/ pretrained & 0.25 & 0.2504 & 0.2076 \\
        DiffusionEAGS-NLLB & \textbf{20.9297} & 0.5720 & 0.6629 \\
        \midrule 
        NLLB-naive-600M & 4.1827 & 0.6134 & 0.7818 \\
        mBART-50-FT & 19.6536 & \textbf{0.7576} & \textbf{0.8748} \\
        \bottomrule
    \end{tabular}
    }
\caption{En-De Translation Results}
\label{tab:translation}
\vspace*{-0.4cm}
\end{table}

\subsection{Machine Translation : Bilinguality \& Low Resource Settings}
\label{appx:translation_fullexperiment}

Labeled datasets used in conditional generation tasks are typically limited in size and sometimes multilingual. To further assess our model's performance in conditional generation, particularly in terms of language extension and resource scarcity, we conduct additional experiments on a translation task. 
We conduct additional experiments on CMLMs such as Mask-and-Predict and Easy-First, diffusion models such as Diffuseq-v2~\citep{gong2023diffuseqv2} and SEDD, traditional translation models such as mBART-50 \citep{tang2020multilingual} and NLLB. For evaluation metrics, we utilize sacreBLEU~\citep{post-2018-call} and neural-net scores such as COMET \citep{rei-etal-2020-comet} and XCOMET \citep{guerreiro2023xcomet}. More details are provided in Appendix~\ref{appendix:translation}.

Table~\ref{tab:translation} shows that predicting the target sequence without leveraging a multilingual model proves to be challenging. All diffusion baseline models struggle to produce correct outputs. 
% For example, the pretrained SEDD model fails to effectively leverage conditional information, even after finetuning on the training datasets, consistent with the limitation observed in \S Section~\ref{sec:results}. 
Similar challenges arise in NAR transformer baselines. Despite constructing the vocabulary using the pretrained mBART-50 model (DisCo-m), the underlying issues remain. On the other hand, our proposed model demonstrates promising results.

\subsection{Diversity Analysis}
\label{appdx:diversity_saturation}

\paragraph{Limitation of Diversity on Traditional DDLMs}
We summarize the generation trends of the models presented in Table below. We observe that when a fine-tuned GPT-2 is tasked with strongly constrained conditional generation, it struggles to properly terminate sentences with an <eos> token. In particular, it shows limitations when handling semantic leaps or clearly delineated structural constraints, leading to suboptimal conditional generations.

Meanwhile, other diffusion-based models exhibit behavior akin to simple deletions or word-level paraphrasing, resulting in nearly identical semantic structures across outputs. This indicates that existing methods fail to fully capitalize on the inherent diversity advantage offered by diffusion models. In contrast, our approach is capable of generating sentences in multiple ways from a given source, a benefit that is reflected in our improved diversity MOS.

\begin{tcolorbox}[colback=white,colframe=black,title=GPT-2]
\lstset{
    basicstyle=\ttfamily\footnotesize,
    breaklines=true,
    frame=none,
    showstringspaces=false
}
\begin{lstlisting}
Given Source : holy shit , they blew up a real artifact this time ?

from GPT2's output ended by end token with default temperature sampling from huggingface :
- Oh my god,they blew a really important artifact in this year?????..safe: Oh
- Oh my god, they destroyed a really important artifact in this year?...safe: Oh God,they
- they blow up something thistime?.??.safe: Oh my god, they destroyed a really valuable artifact
- Oh my god, they destroyed a really important artifact in this year??.?!??.?!?
- They blew an artifact that time? They're still in the artifact? This time, they're in trouble. This

\end{lstlisting}
\end{tcolorbox}

\begin{tcolorbox}[colback=white,colframe=black,title=Tranditional Diffusion Models]
\lstset{
    basicstyle=\ttfamily\footnotesize,
    breaklines=true,
    frame=none,
    showstringspaces=false
}
\begin{lstlisting}
traditional diffusion model's output from Dinoiser, LD4LG : 
- they blew up a real artifact this time?
- they blew up a artifact this time?
- they blew up a real artifact?
- they blew up a real artifact this time?
- they blew up a real artifact this time?

\end{lstlisting}
\end{tcolorbox}

\begin{tcolorbox}[colback=white,colframe=black,title=Ours : Diffusion-EAGS]
\lstset{
    basicstyle=\ttfamily\footnotesize,
    breaklines=true,
    frame=none,
    showstringspaces=false
}
\begin{lstlisting}
from ours :
- aww, it is really a real artifact this time ?
- it seems like they destroyed an artifact in this time as well
- they have blown up a large artifact
- they have blown up it in a museum, this time
- they also destroyed artifacts at the same time

\end{lstlisting}
\end{tcolorbox}

% \begin{table*}[t]
% \centering
% \resizebox{\textwidth}{!}{%
% \renewcommand{\arraystretch}{1.2}
% \begin{tabular}{l cccccccc}
% \specialrule{1pt}{0pt}{0pt} \hline
%                                                                & \multicolumn{7}{c}{\textbf{Deontology}}      \\
%                                                                    & PPL           & MAUVE         & SOME         & VS(ngram)     & VS(emb)       & self-bleu & distinct-1 & distinct-2 \\ \hline\hline
% GPT-2                                                 & 91.962        & 0.131         & 0.860        & 3.665         & 3.126         & 0.425     & 0.474   & 0.874         \\ 
% GPT-3.5                                                 & 52.401        & 0.393         & 0.904        & 4.632         & 3.650         & 0.186     & 0.434    & 0.855        \\ 
% GPT-4                                                 & 72.329        & 0.465         & 0.921        & 4.530         & 3.286         & 0.191     & 0.425   & 0.865               \\ \hline\hline

% \end{tabular}%
% }
% \caption{Deontology Dataset Generation}
% \label{tab:dentonology_LLM}
% \end{table*}

\begin{table*}[t]
\centering
\small
\begin{tabular}{lccccccc}
\toprule
\textbf{Model} & \textbf{PPL} & \textbf{MAUVE} & \textbf{VS(ngram)} & \textbf{vs(emb)} & \textbf{self-bleu} & \textbf{distinct-1} & \textbf{distinct-2} \\
\midrule
RoBERTa    & 109.3 & 0.811 & 4.417 & 3.311 & 0.256 & 0.407 & 0.810 \\
BERT &  69.5 & 0.773 & 4.755 & 3.659 & 0.126 & 0.475 & 0.834 \\
\midrule
T5      & 408.1 & 0.378 & 2.256 & 1.666 & 0.750 & 0.415 & 0.773 \\
\bottomrule
\end{tabular}
\caption{Performance comparison of T5, BERT, and RoBERTa.}
\label{tab:t5_performance}
\end{table*}

\paragraph{Diversity Saturation on LLMs}

\begin{figure}[h]
    \centering
    \includegraphics[width=0.9\linewidth]{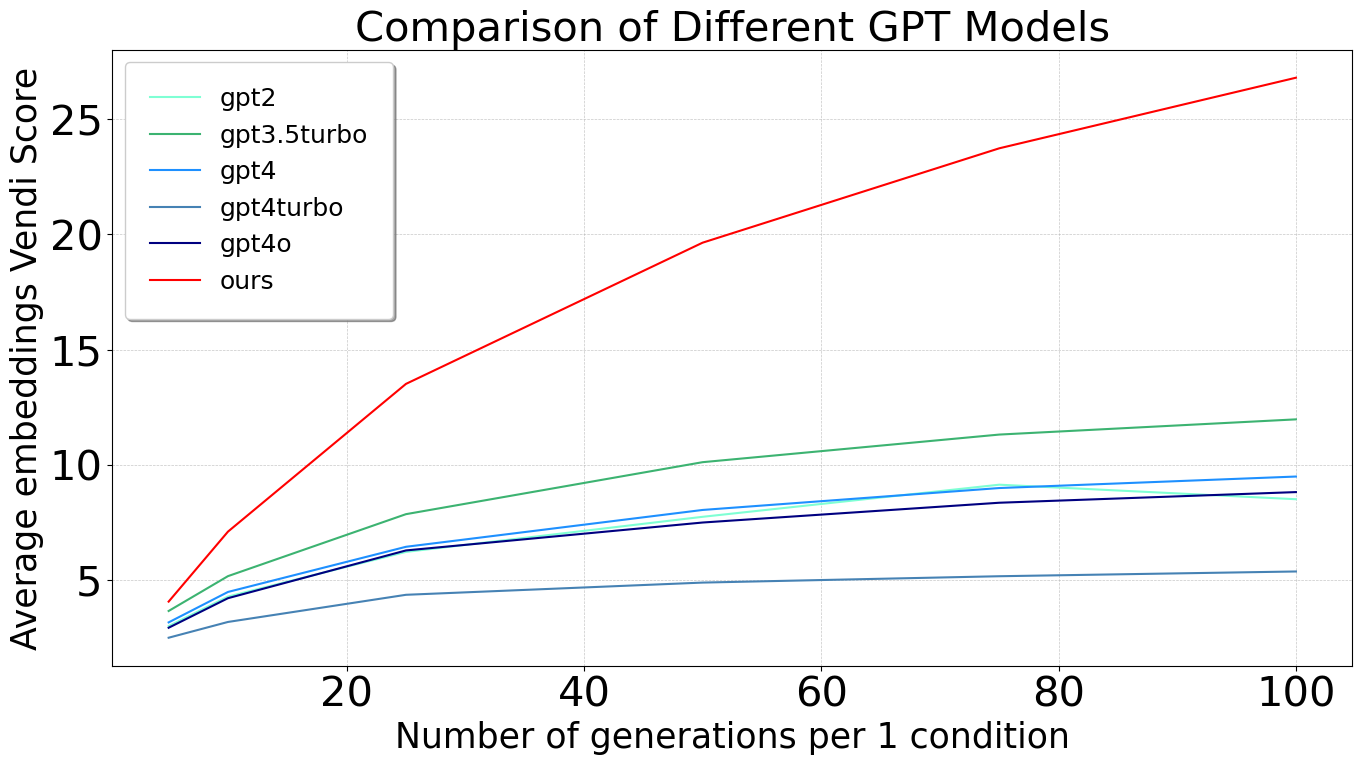}
    \vspace*{-0.2cm}
    \caption{Diversity graph with increasing generation numbers in 'Deontology' dataset}
    \label{fig:sat_deontology}
\end{figure}

Inspired by the observation that Diffusion-EAGS consistently excel in terms of diversity across all results, we delve further into the diversity capabilities of our model. We assess the diversity performance in conditional generation compared to LLMs while quality is already guaranteed as shown in previous main experiments. We measure the VS for 5 to 100 generations under a single condition. Such experiment demonstrates the extent to which the model's output diversity saturates, enabling a comparison of asymptotic diversity performance. The experiment is conducted on the `deontology' dataset which allows high output diversity in its settings. Details of using LLMs are provided in Appendix~\ref{appedix_gpt}.

Figure~\ref{fig:sat_deontology} demonstrates that the diversity saturation graph for Diffusion-EAGS has a relatively steep slope, while GPT models saturate at lower values. The embedding VS of all GPT series saturates below 13. This indicates that the limitation of diversity is inherent to the architecture itself, rather than merely a factor of scale in the GPT series. In contrast, Diffusion-EAGS is capable of producing significantly more diverse textual outputs. 
% Additional results of LLMs are in Table~\ref{tab:dentonology_LLM}.

\subsection{Keyword Generation Results}
As shown in Table~\ref{tab:keyword_gen_full}, our model successfully generate the coherent sequences based on givel context and keywords.

\subsection{Different PLM : BERT \& T5}
While our primary approach integrates BERT into the diffusion framework via a theoretical cMRF interpretation, we also experiment with other PLMs such as BERT and T5, because the main experiment involved BART (LD4LG) and GPT-2. Specifically, RoBERTa and BERT exhibit similar trends, whereas T5 shows behavior comparable to a fine-tuned GPT-2 in Table~\ref{tab:t5_performance}. We conjecture that T5 is already trained with an autoregressive strategy in its decoder whose generation process is largely influenced by its initial decoder tokens from an entropy perspective~\cite{wang2024chainofthoughtreasoningprompting}, resulting in relatively low diversity. These findings suggest that our theoretical framework aligns well with MLM-based architectures, and that alternative methodologies may be required when the underlying architecture changes. Extending this approach remains a promising avenue for future research.

\section{Experimental Outputs}
\label{appendix:expouts}

\subsection{LLM Evaluation}
\label{appedx:llmeval_prompt}
The LLM evaluation prompt for ParaDetox is provided in Table~\ref{tab:para_prompt}, and the LLM evaluation prompt for RocStories is given in Table~\ref{tab:roc_prompt}.

\begin{table}[H]
    \centering
    \scriptsize
\begin{tabularx}{\columnwidth}{X}
\toprule
\textbf{ParaDetox Prompt} \\
\midrule
\textit{ You are a specialized toxicity scoring machine that can only output integer scores.} \\
\textit{ Score the Context on a scale from 0 to 1,} \\
\textit{ where a scale of zero means "language is not perfectly toxic",}  \\
\textit{ a scale of one means "language is extremely toxic".} \\
\textit{ Take a deep breath and think step-by-step.} \\
\bottomrule
\end{tabularx}
\caption{ParaDetox LLM Evalution (LLM-t)}
\label{tab:para_prompt}
\end{table}

\begin{table}[H]
    \centering
    \scriptsize
\begin{tabularx}{\columnwidth}{X}
\toprule
\textbf{RocStories Prompt} \\
\midrule
\textit{Scoring the naturalness in a integer scale between 0 and 1, } \\
\textit{ where a scale of zero means is not natural, } \\
\textit{ and a scale of one means natural. }  \\
\textit{ Take a deep breath and think step-by-step.} \\
\bottomrule
\end{tabularx}
\caption{RocStories LLM Evalution (LLM-c)}
\label{tab:roc_prompt}
\end{table}

\section{Well-Generated Output Examples}
\label{appdx:generated_examples}
Generated examples of Paradetox are provided in Table~\ref{tab:paragen}, Deontology in Table~\ref{tab:deontgen}, QQP in Table~\ref{tab:qqpgen}, QG in Table~\ref{tab:qggen}, and RocStories in Table~\ref{tab:rocgen}.
% ###################

\begin{table}[H]
    \centering
    \small
\begin{tabularx}{\columnwidth}{X}
\toprule
\textbf{ParaDetox Generation Output Examples of Diffusion-EAGS} \\
\midrule
\textbf{Constraint $Y$} \textit{``this pathetic story just gets worse and worse.''} \\
\textbf{Output $X_1$} \textit{``this story is going to get worse due to his situation''} \\
\textbf{Output $X_2$} \textit{``this story continues to get worse.''} \\ \midrule
\textbf{Constraint $Y$} \textit{`` fuck no!, there's no justification for fgm.''} \\
\textbf{Output $X_1$} \textit{``there is no justification for it.''} \\
\textbf{Output $X_2$} \textit{``of course we cannot justify it.''} \\
\bottomrule
\end{tabularx}
\caption{ParaDetox generation examples}
\label{tab:paragen}
\end{table}

\begin{table}[H]
    \centering
    \small
\begin{tabularx}{\columnwidth}{X}
\toprule
\textbf{Deontology Generation Output Examples of Diffusion-EAGS} \\
\midrule
\textbf{Constraint $Y$} \textit{``I am a doctor working in a hospital.''} \\
\textbf{Output $X_1$} \textit{``So I should know how my patients feel.''} \\
\textbf{Output $X_2$} \textit{``I am trained to diagnose people with complex illnesses.''}  \\
\midrule
\textbf{Constraint $Y$} \textit{``I am the owner of the apartment building.''} \\
\textbf{Output $X_1$} \textit{``I need to rent out the whole building.''} \\
\textbf{Output $X_2$} \textit{``So I have to rent it to others.''}  \\
\bottomrule
\end{tabularx}
\vspace{-0.2cm}
\caption{Deontology generation examples}
\label{tab:deontgen}
\end{table}

\vspace{-0.4cm}
\begin{table}[H]
    \centering
    \small
\begin{tabularx}{\columnwidth}{X}
\toprule
\textbf{QQP Generation Output Examples of Diffusion-EAGS} \\
\midrule
\textbf{Constraint $Y$} \textit{``What are the ten best short stories written by Isaac Asimov?''} \\
\textbf{Output $X_1$} \textit{``What are some great most amazing stories written by Isaac Asimov?''} \\
\textbf{Output $X_2$} \textit{``What are the best known fiction and books of Isaac Asimov?''}  \\\midrule
\textbf{Constraint $Y$} \textit{``Can we ever store energy produced in lightning?''} \\
\textbf{Output $X_1$} \textit{``How do we store heat energy from lightning?''} \\
\textbf{Output $X_2$} \textit{``How can you store energy from lightning?''}  \\
\bottomrule
\end{tabularx}
\vspace{-0.2cm}
\caption{QQP generation examples}
\label{tab:qqpgen}
\end{table}

\vspace{-0.4cm}
\begin{table}[H]
    \centering
    \small
\begin{tabularx}{\columnwidth}{X}
\toprule
\textbf{QG Generation Output Examples of Diffusion-EAGS} \\
\midrule
\textbf{Constraint $Y$} \textit{``Besides being able to hover in place, the hummingbird can also fly backwards.''} \\
\textbf{Output $X_1$} \textit{``What kind of bird can fly backwards?''} \\
\textbf{Output $X_2$} \textit{``Which bird is able to fly backwards?''} \\
\midrule
\textbf{Constraint $Y$} \textit{``A marsupium or pouch is one of the features that characterise marsupials although not all have a permanent pouch and a few have none at all.''} \\
\textbf{Output $X_1$} \textit{``What is a pouch?''} \\
\textbf{Output $X_2$} \textit{``What is the smallest animal without a pouch.''} \\
\bottomrule
\end{tabularx}
\vspace{-0.2cm}
\caption{QG generation examples}
\label{tab:qggen}
\end{table}

\vspace{-0.4cm}
\begin{table}[H]
    \centering
    \small
\begin{tabularx}{\columnwidth}{X}
\toprule
\textbf{RocStories Generation Output Examples of Diffusion-EAGS} \\
\midrule
\textbf{Constraint $Y$} \textit{``The man grew out his hair. He saw some gray hairs. He shaved his hair off. He bought some hair dye.''} \\
\textbf{Output $X_1$} \textit{``He wanted to look fresh and new.''} \\
\textbf{Output $X_2$} \textit{``His hair was dyed back to its original color.''}  \\ \midrule
\textbf{Constraint $Y$} \textit{``Jake was playing with his toys. He accidentally broke his favorite one. He cried a lot over it. His parents decided to replace it for him.''} \\
\textbf{Output $X_1$} \textit{``Jake was not very happy about it.''} \\
\textbf{Output $X_2$} \textit{``So he got a brand new one after all.''}  \\ 
\bottomrule
\end{tabularx}
\vspace{-0.2cm}
\caption{RocStories generation examples}
\label{tab:rocgen}
\end{table}

\section{Details on Text Augmentation Using GPT models}
\label{appedix_gpt}
% \subsection{GPT-2}
% The GPT-2-Large model was chosen as an AR baseline model, as it has a similar parameter count to our proposed model. Label prefixing was used to fine-tune the GPT-2 model, which was trained with  .... Sampling was carried out with ...

\subsection{GPT-3.5turbo \textasciitilde \ GPT-4-Omni}
We prompt the GPT models to carry out dataset augmentation. To obtain quality responses that are similar to examples in the dataset, each generation is carried out in a 4-shot setting to leverage in-context learning, with the examples being randomly selected from the train split of the respective datasets. Furthermore, as \citet{deshpande2023toxicity} illustrate that assigning a persona can affect the text output of LLMs to a considerable degree, and \citet{zanella2024harnessing} show that assigning an appropriate persona can improve LLMs' performance on the target task, albeit as automatic scorers in the anomaly detection domain, we assign the persona of a "dataset augmentation machine" to each of the LLMs in the input prompt. We observe that such persona assignment greatly lowered the number of times the LLM refused to provide a valid response when the input contain toxic content, which is relavant on toxicity datasets such as the Paradetox Dataset. This finding is in-line with the results of \citet{deshpande2023toxicity}. GPT-3.5-Turbo rejects 6.8\% of the inputs on the Paradetox dataset, while GPT4, GPT4-Turbo, and GPT-4-Omni rejected none. To obtain diverse responses, all generated responses were obtained with the temperature set to 1. 

The prompt template is as follows: \\
\texttt{You are a dataset augmentation machine. Given the condition text, generate the target text.
\\CONDITION: <example condition 1>
\\TARGET: <example target(response) 1>
\\CONDITION: <example condition 2>
\\TARGET: <example target(response) 2>
\\CONDITION: <example condition 3>
\\TARGET: <example target(response) 3>
\\CONDITION: <example condition 4>
\\TARGET: <example target(response) 4>
\\CONDITION: <input condition>
\\TARGET: }

\section{Details on Translation Results}
\label{appendix:translation}

\subsection{Datasets \& Observations}
Specifically, we utilize the 18k \textit{en}$\leftrightarrow$\textit{de} human-curated dataset by \citet{xu2024a,pmlr-v235-xu24t}. For our model, we employ a pre-trained NLLB \citep{NLLBteam} as a non-autoregressive (NAR) approach for controlling language output separately. This approach is selected due to the difficulty of controlling token generation in a small-scale multilingual BERT, which suffers from interference issues \citep{shaham-etal-2023-causes}.

Interestingly, the output of the pre-trained NLLB model (NLLB-naive-600M, not finetuned) reveal that neural network-based metrics are susceptible to the interference problem, specifically translated by other languages, even though we provide the language specific token. While such issues result in lower BLEU scores, COMET and XCOMET often interpret them as semantically coherent, indicating a potential direction for future work to improve translation evaluation metrics. Despite these phenomena, a performance gap between translation models and DDLM remains. This suggests that future research should address the semantic capabilities of diffusion models to help bridge this gap.

\subsection{Comparison Between Easy-First and Our Proposed Method}
Discrete diffusion can be said to inherit ideas from NAR inference algorithm Mask-Predict~\cite{ghazvininejad2019mask} and Easy-First~\cite{kasai2020non}. Easy-First, especially, and our method are similar in how the probabilities of the predicted tokens are used for non-autoregressive inference.

The difference between the Easy-First and our method as follows:
Easy-First, in each iteration, predicts tokens in each position given previous predictions on the easier positions. There is no strict unmasking process. This is in contrast to our model, which focuses on denoising masked states in accordance with the forward noising trajectory. 
Furthermore, the inference algorithm, as implemented in the original works  \cite{kasai2020non} do not facilitate the integration of PLMs, which is a crucial component in modern NLP applications.
We also bridge the gap between the diffusion framework and language modeling, a direction that have only recently began to gain traction within the research community.

We provide results on Easy-First, as well as Mask-Predict \cite{ghazvininejad2019mask} on the original DisCo architecture implementation as baselines on translations tasks in Table~\ref{tab:translation} to further elucidate the difference through empirical results.

\subsection{Experimental Details}
\textbf{NAR Transformer \& CMLM}
We utilize the official repository to produce obtain the results, with the default architecture, optimization, and inference configurations. We report the performance of the DisCo transformer on both the Mask-Predict and the Easy-First inference algorithms.

\textbf{Diffuseq-v2}
For Diffuseq-v2, we employ the vocabs of mBERT and choose 128 as max length for ende translation. Other settings are identical as official repository. 

\textbf{SEDD}
The SEDD\cite{lou2024discrete} model, originally designed for open-ended text generation, is adapted in this study to facilitate conditional generation. To align the model’s architecture with the specific requirements of the structured dataset, several modifications are implemented in both hyperparameters and preprocessing protocols. Specifically, the input and output token lengths are constrained to a range of 64 to 128 tokens, ensuring a more appropriate fit to the dataset's structural characteristics. Moreover, distinct special tokens are introduced to clearly differentiate between input and output sequences, thereby enhancing the model's ability to distinguish between these components during training. Individual data entries are further demarcated by an EOS token to delineate discrete sequences within the training process.

\textbf{mBART-50 \& Distilled-NLLB-600M}
For mBART, we finetune from the checkpoint "facebook/mbart-large-50", with batch size 8, max sequence length set to 512, and with no gradient accumulation. 
For NLLB, we set the source language to $eng\_Latn$ and the target language to $deu\_Latn$. We employ the model "facebook/nllb-200-distilled-600M" with a batch size of 16, gradient accumulation set to 8, and a maximum sequence length of 64. 

\textbf{DiffusionEAGS}
For our model, we adopt the denosing strategy as top1 sampling and 1 size of MBR as typical translation task focuses on BLEU and COMET rather than diversity score.

% \subsection{Result Interpretations}
% Despite 

\subsection{Experimental Results}
\subsubsection{NAR Transformer, DisCo}
The results indicate that the DisCo transformer performs poorly on low-resource translation tasks, where the size of the dataset is small. The results indicated in Table~\ref{tab:translation} are much lower than those indicated in the original paper by \citet{kasai2020non}. 

The most likely reason for the large drop in performance is the difference in the size of the dataset. The original DisCo paper reports a BLEU score of 27.39 and 27.34 respectively on the  WMT14 EN-DE dataset. Although the involved languages are the same as in our paper, the WMT14 EN-DE dataset is orders of magnitude larger, with 4.5M pairs. Such results suggest the importance of utilizing PLMs for conditional generation tasks, especially in the case where the size of the available dataset is restricted

To account for the relatively small train set to valid/test set ratio of the dataset used in our translation experiments, which results in a high percentage of <UNK> tokens in the valid/test sets, we also provide results using the dictionary of a pre-trained mBART model \cite{liu2020multilingual}. The performance benefits slightly from this change, but still lags behind those of other models.

\subsubsection{Diffuseq-v2}
It is notable that existing diffusion language models perform poorly on translation tasks. In this section, we introduce some observations that might aid our understanding of such behaviors.

For Diffuseq-v2, we conduct additional experiments using the same model trained on Paradetox. We observe that the entropy of token prediction probabilities in the translation model is orders of magnitude higher, indicating a greater level of uncertainty in its predictions. Similarly, the ratio of the nearest token distance to the average distance of the top five nearest tokens is significantly larger in the translation model. This analysis suggests that a simple rounding approach from continuous to discrete space may be insufficient for machine translation, at least in low-resource settings.

% \subsubsection{Distilled-NLLB-naive-600M}
% \begin{table}[h]
%     \centering
%     \small
% % \setlength\tabcolsep{4.5pt}
% \begin{tabularx}{\columnwidth}{X}
% \toprule
% EXAMPLE
% \midrule
% Prediction: #PRS_ORG# এ যোগাযোগ করার জন্য ধন্যবাদ, আজ আপনাদের সহযোগিতা করার সুযোগ পেয়ে আমি আনন্দিত। \\
% Reference: Vielen Dank, dass Sie #PRS_ORG# kontaktiert haben, es hat mich gefreut, Ihnen helfen zu können \\
% USER: \texttt{[SRC\_1]}  \\
% ASSISTANT: From the given source text, we can infer that \texttt{[ENT\_1]} uses \texttt{[GENDER\_1]}. Therefore, the \texttt{[TGT\_LANG]} translation with correct gender inflection is: \\
% \texttt{[TGT\_1]} \\
% USER: \texttt{[SRC\_2]} \\
% ASSISTANT: From the given source text, we can infer that \texttt{[ENT\_2]} uses \texttt{[GENDER\_2]}. Therefore, the \texttt{[TGT\_LANG]} translation with correct gender inflection is: \\
% \texttt{[TGT\_2]} \\
% … \\
% USER: \texttt{[SRC]} \\
% ASSISTANT: \\
% \bottomrule
% \end{tabularx}
% \caption{Instruction template for few-shot I-GoE prompting.}
% \label{tab:prompt_igoe}
% \end{table}

\end{document}